\documentclass{svjour3}   
\usepackage{epsfig}
\usepackage{amstext,amssymb}
\usepackage[usenames,dvipsnames]{color}
\epsfclipon 
\begin{document}
\title{Fast Inference of Interactions in Assemblies of Stochastic
Integrate-and-Fire Neurons from Spike Recordings }
\author{R. Monasson$^{1,3}$, S. Cocco$^{2,3}$}
\institute{
$^1$ Laboratoire de Physique Th\'eorique de l'ENS, CNRS \& UPMC, 24 rue Lhomond,
75005 Paris, France\\
$^2$ Laboratoire de Physique Statistique de l'ENS, CNRS \& UPMC, 24 rue Lhomond,
75005 Paris, France\\
$^3$ The Simons Center for Systems Biology, Institute for Advanced Study, 
Einstein Drive, Princeton NJ 08540, USA}
\date{ }
\maketitle
\begin{abstract}
We present two Bayesian procedures to infer the interactions and
external currents in an assembly of stochastic integrate-and-fire
neurons from the recording of their spiking activity. The first
procedure is based on the exact calculation of the most likely
time courses of the neuron membrane potentials conditioned by the
recorded spikes, and is exact for a vanishing noise variance and for an instantaneous synaptic integration. The second
procedure takes into account the presence of fluctuations around the
most likely time courses of the potentials, and can deal with moderate
noise levels.  The running time of both procedures is proportional to
the number $S$ of spikes multiplied by the squared number $N$ of neurons. The
algorithms are validated on synthetic data generated by networks
with known couplings and currents.  We also reanalyze previously
published recordings of the activity of the salamander retina
(including from 32 to 40 neurons, and from $65,000$ to $170,000$ spikes). We
study the dependence of the inferred interactions on the membrane leaking time; the 
differences and similarities with the classical cross-correlation 
analysis are discussed.
\end{abstract}

\section{Introduction}

\begin{figure}[t]
\begin{center}
\epsfig{file=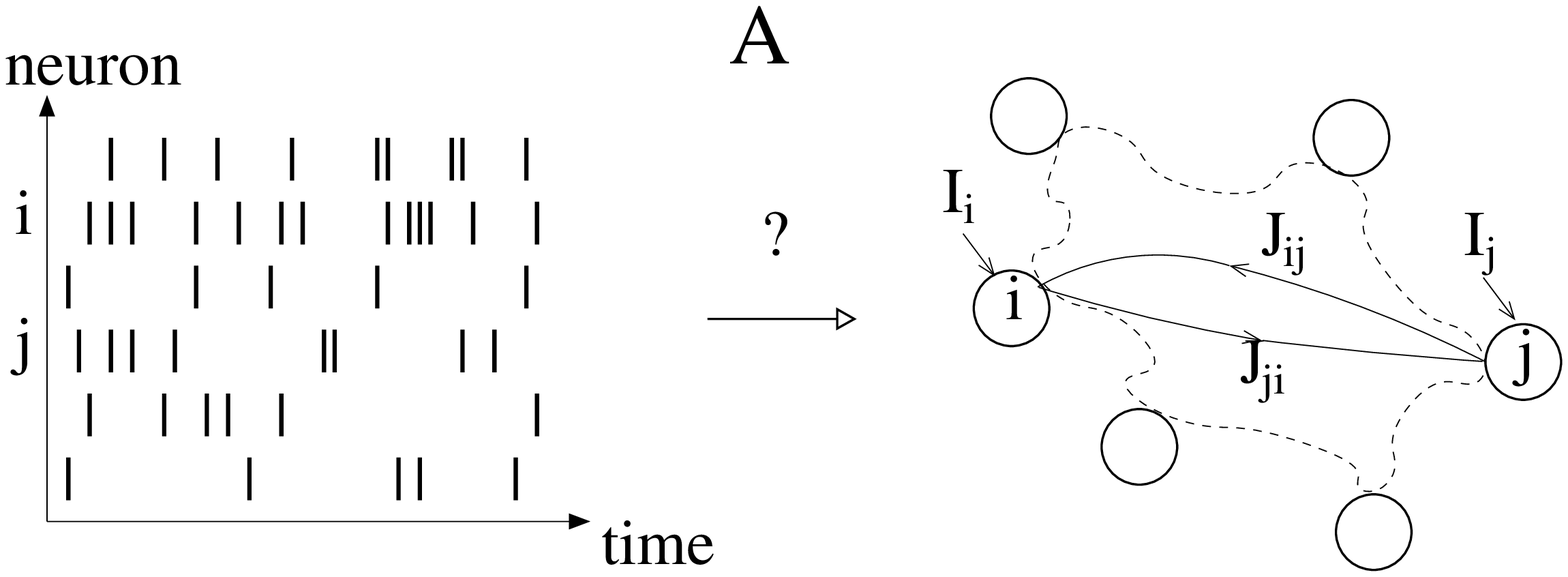,width=6.6cm} \hspace{1cm}
\epsfig{file=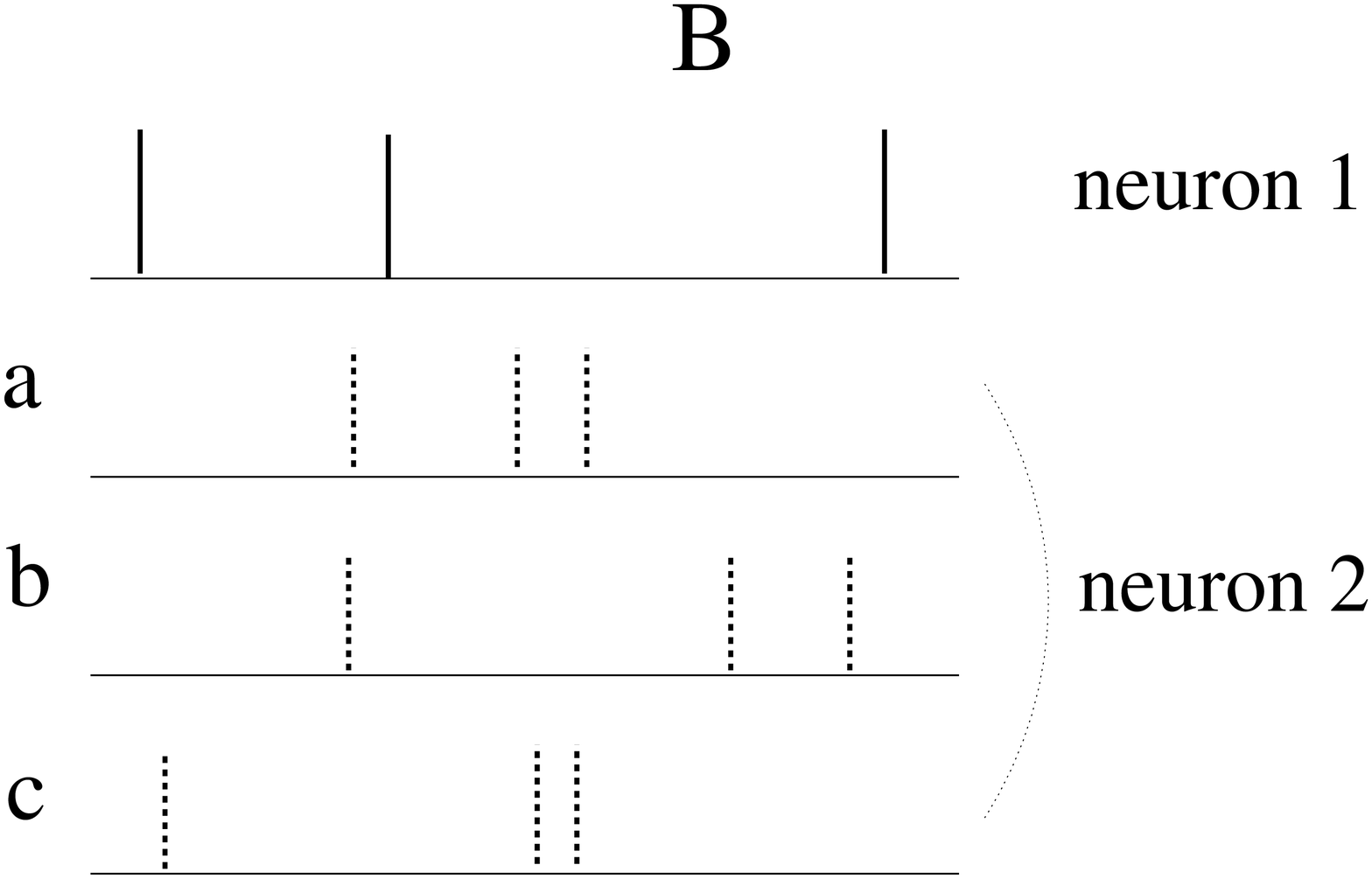,width=4.4cm}
\caption{{\bf A.} Extra-cellular recordings give access, through
spike-sorting, to the times $t_{i,k}$ of the spikes emitted by a
population of neurons. We want to infer the values of the interactions
$J_{ij}$ and external inputs $I_i$ of the network most likely to have
generated the recorded activity.  {\bf B.}~Example of firing activity
of $N=2$ neurons. The top panel shows three spikes emitted by neuron
1. Panels a, b, c show possible activities of neuron 2, with equal
average firing rates but with different timings.}
\label{fig-1}
\end{center}
\end{figure}

Over the past decades, multi-electrode recordings (Taketani and
Baudry, 2006) have unveiled the nature of the activity of populations
of neural cells in various systems, such as the vertebrate retina
(Schnitzer and Meister, 2003), cortical cultures (Tang et al, 2008),
or the prefrontal cortex (Peyrache et al, 2009).  The observation of
substantial correlations in the firing activities of neurons has
raised fundamental issues on their functional role (Romo, Hernandez,
Zainos and Salinas, 2003; Averbeck, Latham and Pouget, 2006). 
From a structural point of view, a challenging problem is
to infer the network and the strengths of the functional interactions between
the neural cells from the spiking activity (Fig.~\ref{fig-1}A). Powerful
inference procedures are needed, capable to handle massive data sets,
with millions of spikes emitted by tens or hundreds of neurons.

A classical approach to infer functional neural connectivity is through
the study of pairwise cross-correlations (Perkel, Gerstein and Moore,
1967; Aersten and Gerstein, 1985).  The approach was applied in a
variety of neural systems, including the auditory midbrain of the
grassfrog (Epping and Eggermont, 1987), the salamander retina
(Brivanlou, Warland and Meister, 1998), the primate and rat prefrontal cortex
(Constantidinidis, Franowicz and Goldman-Raking, 2001; Fujisawa,
Amarasingham, Harrison and Buzsaki, 2008).  Other approaches,
capable of taking into account network-mediated effects, were proposed
based on concepts issued from statistics and graph theory (Seth and Edelman, 2007; 
Dahlhaus, Eichler and Sandk\"uhler, 1997; Sameshima and
Baccal\'a, 1999; Jung, Nam and Lee, 2010), information theory
(Bettencourt, Stephens, Ham and Gross, 2007), or statistical physics
(Schneidman, Berry, Segev and Bialek, 2006; Shlens et al, 2006). 

An alternative approach is to assume a particular dynamical model for
the spike generation. The generalized linear model, which
represents the generation of spikes
as a Poisson process with a time-dependent rate is a popular framework 
(Brown, Nguyen, Frank, Wilson and Solo, 2001; Truccolo et al, 2005; Pillow et al, 2008).  
The Integrate-and-Fire (IF) model, where spikes
are emitted according to the dynamics of the membrane potential is another
natural candidate (Gerstner and Kistler, 2002; Jolivet, Lewis and Gerstner, 2004). 
The problem of estimating the model parameters (external current, variance of the
noise, capacitance and conductance of the membrane) of a
single stochastic IF neuron from the observation of a spike train has
received a lot of attention (Paninski, Pillow and
Simoncelli, 2004; Pillow et al., 2005; Mullowney and Iyengar, 2008;
Lansky and Ditlevsen, 2008).  Few studies have focused on the
inference of interactions in an assembly of IF neurons (Makarov,
Panetsos and de Feo, 2005). Recently, we proposed a Bayesian algorithm
to infer the interactions in a network of stochastic perfect
integrators when the synaptic integration is instantaneous and the noise is vanishingly small (Cocco, Leibler and Monasson, 2009).

In the present work we introduce a Bayesian algorithm to infer the couplings and the external currents in an assembly of leaky IF neurons, and in presence of moderate input noise (Fig. \ref{fig-1}A). The computational time grows as the product of the number of recorded spikes, and the square of the number of neurons. We validate the algorithm
on synthetic data, and apply it to real recordings of the ganglion
cell activity in the salamander retina, presented with natural visual
stimuli, and in the absence of stimulus (spontaneous activity).

\section{Materials and Methods}
\label{defsec1}

\subsection{Definition of the Leaky Integrate-and-Fire model}

In the Leaky Integrate-and-Fire (LIF) model,
the membrane potential $V_i(t)$ 
of neuron $i$ at time $t$ obeys the first-order differential equation,
\begin{equation} \label{ode}
C\frac{dV_i}{dt}(t) = - g\, V_i(t) + I_i^{syn} (t)+ I_i +\eta_i(t)
\end{equation}
where $C$ and $g$ are, respectively, the capacitance and conductance
of the membrane.  The ratio $\tau=C/g$ is the membrane leaking time.
$I_i^{syn}(t)$ is the synaptic current coming from the other neurons and entering the neuron $i$ at time $t$:
\begin{equation}\label{defisyn}
I_i^{syn} (t)= \sum _{j (\ne i)} J_{ij} \; \sum _{k} \delta( t - t_{j,k})
\end{equation}
where $J_{ij}$ is the strength of the connection from neuron $j$ onto neuron $i$ (Figure \ref{fig-1}A); $t_{j,k}$ is the time at which neuron $j$ fires its $k^{th}$ spike. We assume that synaptic inputs are instantaneously integrated, {\em i.e.} that the synaptic integration time is much smaller than all the other time scales, including $\tau$. Our method for inferring the interactions relies on this assumption, and should be modified in the presence of synaptic integration kernels with temporal filtering.  $I_i$ is a constant external current flowing into neuron $i$ (Fig. \ref{fig-1}A), and $\eta_i(t)$ is a fluctuating current, modeled as a Gaussian noise process: $\langle \eta _i
(t)\rangle = 0$, $\langle \eta _i (t) \; \eta _j (t')\rangle =\sigma^2 \; \delta _{ij}\;\delta (t-t')$. The noise standard deviation, $\sigma$, has here the dimension of a current times the square root of a time. An alternative definition would consist in rescaling $\sigma$ with a time dependent factor, {\em e.g.} $\sqrt \tau$; our definition allows us to reach the perfect integrator limit ($\tau\to\infty$) while keeping $\sigma$ fixed.

The neuron $i$ remains silent as long as $V_i$ remains below a threshold potential $V_{th}$. If the threshold is crossed at some time $t_{0}$, {\em i.e.} $V_i(t_0)=V_{th}$, then a spike is emitted, and the potential is reset to its resting value: $V(t_0^+)=0$. The
dynamics then resumes following (\ref{ode}).

\subsection{Likelihood of the spiking times for given interactions and currents}

Let ${\cal J}=\{J_{ij}\}$ and ${\cal I}=\{I_i\}$ denote the sets of,
respectively, the interactions and currents.  Let $t_{i,k} \in [0;T]$
be the time at which neuron $i$ emits its $k^{th}$ spike; $T$ is the
duration of the recording.  How can we infer the interactions and
currents from the observation of the spiking activity? Consider the
raster plots in Fig.~\ref{fig-1}B.  In pattern $a$, the timings of the
spikes of neuron 1 do not seem to be correlated to the activity of
neuron 2. Hence, we may guess that there is no interaction from neuron 2
to neuron 1 ($J_{12}=0$). In pattern $b$, a
spike of neuron 1 is likely to follow a spike of neuron 2, which
suggests that the interaction $J_{12}$ is positive. Conversely, in
pattern $c$, it seems that the firing of neuron 2 hinders the firing
of neuron 1, which indicates that $J_{12}$ has a negative value.

This crude reasoning can be made mathematically rigorous in the
framework of statistical inference (Cover and Thomas, 2006).  Let us
define the likelihood $P({\cal T}|{\cal J},{\cal I})$ of a set of
spiking times, ${\cal T}=\{t_{i,k}\}$, given ${\cal J}$ and ${\cal
I}$.  According to Bayes rule the most likely couplings and currents, $\hat
{\cal J}$ and $\hat {\cal I}$, given the set of spiking times ${\cal
T}$ can be inferred through the maximization of $P({\cal T}|{\cal J},{\cal I})$ 
\footnote{We
consider here that the {\em a priori} measure over the couplings and
currents is flat.}.  Due to the statistical independence of the noises
$\eta_i$ from neuron to neuron, the likelihood $P$ of ${\cal T}$ given
${\cal J},{\cal I}$ can be written as the product of First-Passage
Time (FPT) probabilities,
\begin{equation}\label{prodfpt}
P({\cal T}|{\cal J},{\cal I})= \prod_{i,k} p_{FPT} (t_{i,k+1} | t_{i,k}
, \{t_{j,\ell} \}, \{J _{ij}\}, I_i ) \ .
\end{equation}
Here $p_{FPT}$ denotes the probability that $V_i$ crosses $V_{th}$ for 
the first time at time $t_{i,k+1}$, starting from $0$ at time $t_{i,k}$ and 
conditioned to the inputs from the other neurons at times $t_{j,\ell}$,
with $t_{i,k}< t_{j,\ell}< t_{i,k+1}$. As the synaptic integration is 
instantaneous, an incoming spike from neuron $j$ results in a (positive or negative)
jump of the potential $V_i$ by $J_{ij}/C$; 
$p_{FPT}$ can therefore be interpreted as the FPT density probability
for a one-dimensional Ornstein-Uhlenbeck process with a time-dependent force.
It is important to stress that the presence of the products 
over the spike intervals in (\ref{prodfpt}) 
does not entail that the spiking times are independent. 

Consider now the potential $V_i(t)$ during the inter-spike interval 
(ISI) $[t_{i,k};t_{i,k+1}]$.
The boundary conditions are $V_i(t_{i,k}^+)=0$ (reset of the potential
right after a spike), and $V_i(t_{i,k+1}^-)=V_{th}$ (condition for
firing). At intermediate times, the potential can take any value smaller
than $V_{th}$. The logarithm of the probability of a dynamical path
(time course)
of the potential over the $k^{th}$ ISI of neuron $i$ is, after multiplication
by the variance $\sigma^2$ of the noise,
\begin{eqnarray}\label{pathint2}
{\cal L}[ V_i(t); k,{\cal T},{\cal J},{\cal I}]&=& -\frac 1{2}
\int _{t_{i,k}}^{t_{i,k+1}} dt\ \eta_i(t)^2 \\&=& -\frac 1{2}
\int _{t_{i,k}}^{t_{i,k+1}} dt\ \left[ C\,\frac{d V_i}{dt}(t) +g\, V_i(t)
-I_i^{syn} (t)-I_i\right]^2 \ ,\nonumber
\end{eqnarray}
according to the Gaussian nature of the noise $\eta_i(t)$ and to the dynamical
equation of the LIF (\ref{ode}).

\subsection{Dynamical equations for the optimal potential and noise}

While no exact expression is known for $p_{FPT}$, it can
be analytically approximated by the contribution of the most probable 
dynamical path for the potential, $V_i ^*(t)$ (Paninski, 2006). 
This approximation becomes exact when the standard deviation $\sigma$ of the 
noise is small. The idea is to replace 
the distribution of paths for the potential $V_i(t)$  
with a single, most likely path $V_i ^*(t)$, which we call optimal.
We now explain how to derive $V_i^*(t)$ through the condition that
the log--probability ${\cal L}$ (\ref{pathint2}) is maximal.

Let us assume first that $V_i^*(t)<V_{th}$. Then, the derivative of
 ${\cal L}$ in (\ref{pathint2}) with respect to $V_i^*(t)$  must vanish, 
which gives 
\begin{equation}\label{optima}
\left.\frac{\delta {\cal L}}{\delta V_i(t)}\right|_{V_i^*}
= -C^2 \frac{d^2V^*_i}{dt^2}(t)
+g^2\, V_i^*(t) + C\frac{dI_i^{syn}}{dt}(t) -g \, I_i^{syn}(t)-g\, I_i=0
\ .
\end{equation} 
We now turn this second order differential equation for the 
optimal potential into a first order differential equation at the
price of introducing a new function, $\eta^*_i(t)$, and a new first 
order differential equation for this function. It is 
straightforward to check that the solution of
\begin{equation}\label{optimb}
C\frac{dV_i^*}{dt}(t)= - g\; V_i^*(t) + I_i^{syn}(t)+I_i+\eta_i^*(t)
\end{equation} 
is a solution of the optimization equation (\ref{optima}) if
$\eta_i^*(t)$ fulfills 
\begin{equation}\label{2.8} 
\frac{d\eta^*_i}{dt}(t)= \frac gC\; \eta_i^*(t) =\frac{\eta_i^*(t)}\tau\ ,
\end{equation}
where $\tau$ is the membrane leaking time.
The similarity between eqns (\ref{ode}) and (\ref{optimb}) allows us to
interpret $\eta _i^*(t)$ as a current noise. However, this
noise is no longer stochastic, but rather it follows the deterministic path
solution of (\ref{2.8}). We will, therefore, in the following 
refer to $\eta_i^*(t)$ as the optimal noise. 
$\eta_i^*(t)$ corresponds to the most likely value the noise takes given the set
of spiking times. Solving (\ref{2.8}) shows that 
the optimal noise is an exponential function of the time:
\begin{equation}
\eta _i^*(t) = \eta\; \exp( +t/\tau) \qquad (V_i^*(t)<V_{th})
\ , \label{eq1} 
\end{equation}
where $\eta$ is a constant, which we call noise coefficient.

It may happen that the optimal potential only reaches the threshold
without actually crossing it at intermediate times. When this is the
case, the optimal potential equals $V_i^*(t)=V_{th}$ and its derivative
with respect to the time vanishes. The expression for
the optimal noise can be then read from (\ref{optimb}), 
and is given by 
\begin{equation} \label{optimc}
\eta _i^*(t) = g\,V_{th} - I_i^{syn} (t) - I_i \qquad (V_i^*(t) = V_{th})\ . \label{eq2}
\end{equation}
Equation (\ref{eq2}) ensures that the potential
does not cross the threshold value at a time $t < t_{i,k+1}$.

Despite their apparent simplicity, eqns (\ref{optimb},\ref{eq1},\ref{eq2}) 
are not easy to solve, due mainly to the interplay between the two regimes,
$V^*< V_{th}$ and $V^*=V_{th}$, mentioned above. 
The determination of $V_i^*(t)$ was achieved numerically by Paninski for
a single neuron (Paninski, 2006). We now sketch 
the procedure to determine $V_i^*(t)$ rapidly, even for tens of neurons. 
The procedure relies on the search for contacts, that is, 
times at which the optimal potential touches the threshold. There are
two types of contacts: contacts coinciding with a synaptic input
(the potential touches the threshold at time $t_{j,k}$), and contacts arising in between two inputs. In the absence of leakage, only the former type of contacts
matter, and a search procedure to locate those isolated-time contacts 
was proposed by Cocco, Leibler and Monasson (2009). In the presence of
leakage, both types of contacts have to be taken into account. The search procedure
is more complex, and is explained below. 

\subsection{Fixed Threshold procedure: optimal paths for the potential and the noise}
\label{sec-algo}

\begin{figure}
\begin{center}
\epsfig{file=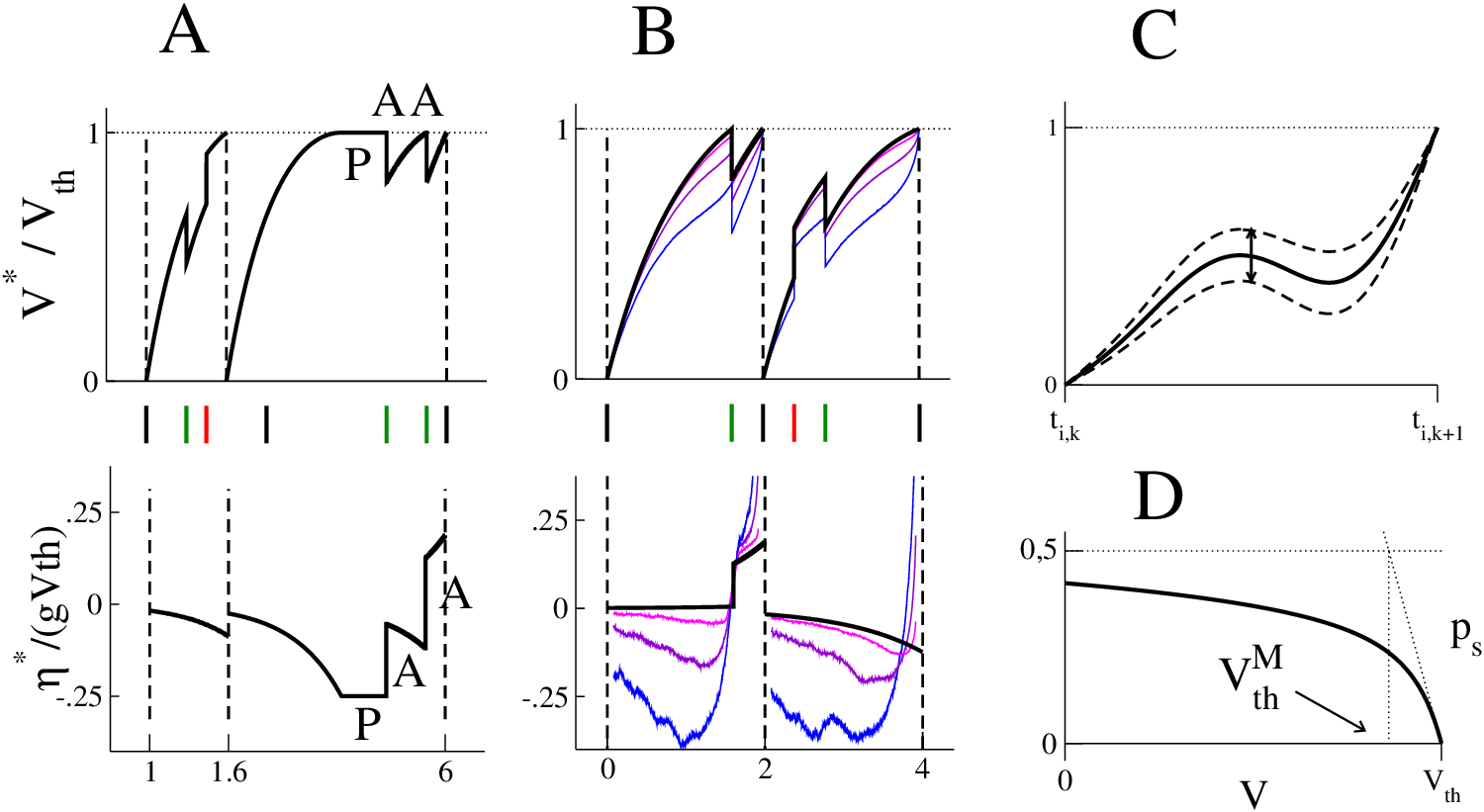,width=7.5cm}
\epsfig{file=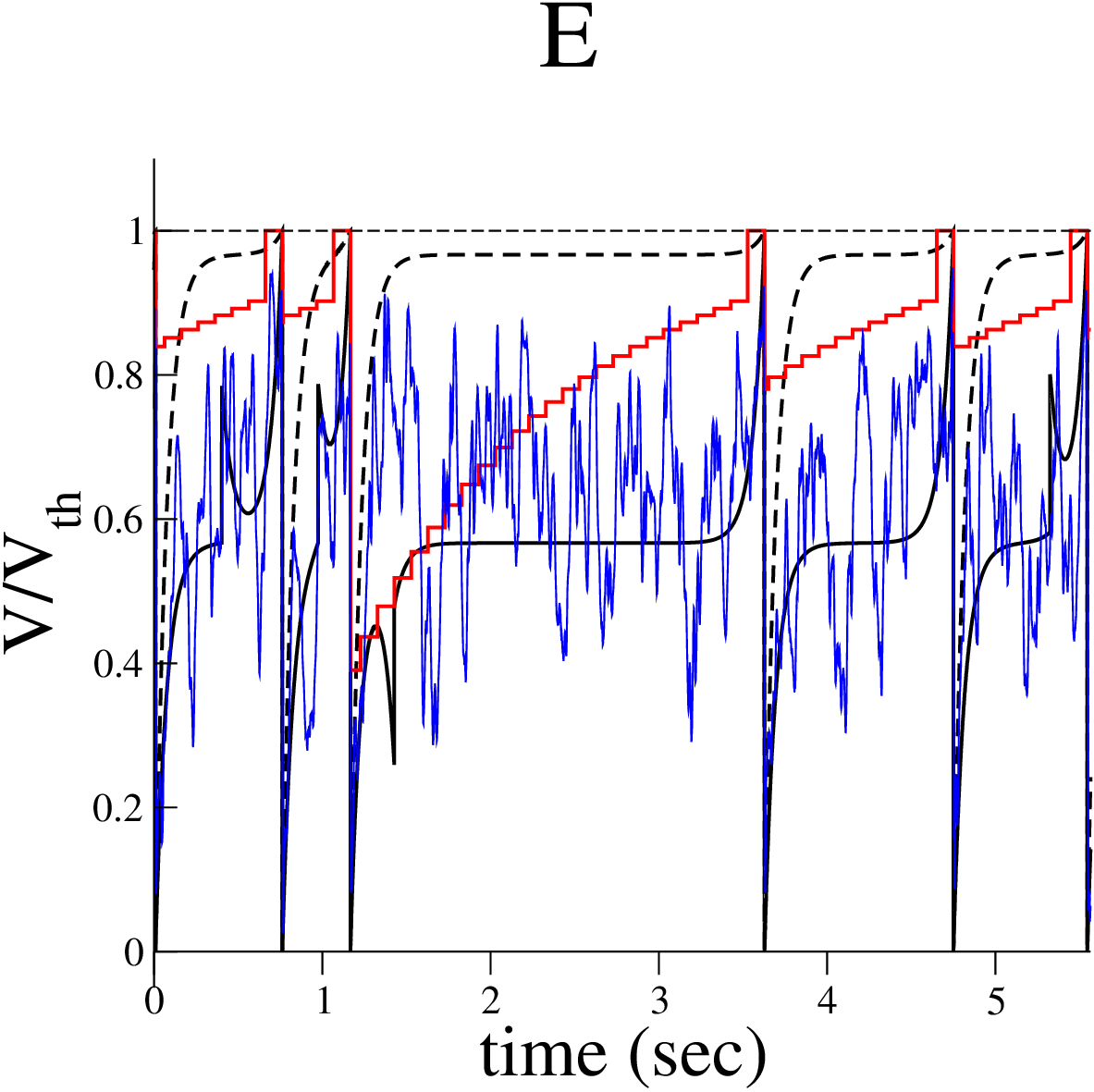,width=4.cm}
\caption{{\bf A \& B.} Sketches of the optimal potentials $V^*$ (top, black curves) and noises $\eta^*$ (bottom, black curves) for one neuron receiving several inputs from two other neurons (\textcolor{red}{red} and \textcolor{OliveGreen}{green} impulses, middle, with $\textcolor{red}{J}_{\textcolor{red}{red}}= -\textcolor{OliveGreen}{J}_{\textcolor{OliveGreen}{green}}=.2\, CV_{th}$). The membrane conductance is $g=.8 I/V_{th}$. The jump in the optimal noise consecutive to an active contact is always positive. {\bf A.} Illustration of Passive (P) and Active (A) contacts.
{\bf B.} Comparison with numerical simulations, averaged over
$\sim 5,000$ samples, for $\bar \sigma=\textcolor{magenta}{.07}$ (red),
$\textcolor{Violet}{.18}$ (purple), $\textcolor{blue}{.36}$ (blue);
noise curves are averaged over the time-window $\Delta t=.015\,\tau$.
{\bf C.} Dashed lines represent possible paths for the potential when the noise 
standard deviation, $\sigma$, does not vanish. The amplitude of the
fluctuations of the potential around $V^*$ at the mid-point of the ISI is symbolized by the 
double arrow line.
{\bf D.}~Probability $p_s(\delta t|V)$ that an Ornstein-Uhlenbeck process
starting from $V$ does not cross the threshold $V_{th}$ for a time 
$\delta t=3.5 \;\tau$. Parameters are $g V_{th}/I=1.2$, $\bar \sigma=.15$. 
The tangent line to $p_s$ in $V=V_{th}$ crosses the $p_s=\frac 12$ line in $V_{th}^M$.
{\bf E.} System of two IF neurons, with $gV_{th}/I_{1}=1.5,gV_{th}/I_{2}=2.,J_{12}/(CV_{th})=.1,J_{21}=0,\bar\sigma=.25$. The dashed and full black curves represent the optimal potentials for neuron 1 calculated by, respectively, the Fixed and Moving Threshold procedures; for the latter, $V_{th}^M$ is shown in red. One random realization of the membrane potential (averaged over a $10$ msec time--window) is shown for comparison (blue curve). }
\label{fig-pro}
\label{fig-fluctuphi}
\label{fig-compar-optpath}
\label{fig-scheme}
\end{center}
\end{figure}

We assume in this Section that the couplings and currents are known. Consider neuron $i$ at time $t\in [t_{i,k};t_{i,k+1}]$, where $k$ is the index of the ISI. The initial
and final conditions for the optimal potential are: $V_i^*(t_{i,k}^+)=0$ and $V_i^*(t_{i,k+1}^-)=V_{th}$. In between,
$V^*_i(t)$ obeys the LIF evolution equation (\ref{optimb}) with an
optimal `noise' $\eta^*_i(t)$. $\eta_i^*(t)$ can be interpreted as a
non-stochastic, external, time-dependent current to be fed into the
neuron in order to drive its potential from 0 to $V_{th}$, given the
synaptic couplings.  The expressions for the optimal `noise' are
given by (\ref{eq1}) when $V^*_i(t)<V_{th}$, and (\ref{eq2}) when the
optimal potential $V_i^*(t)$ is equal to the threshold value. 

When $V_i^*(t)$ reaches the threshold at a time coinciding with an incoming
spike, the coefficient $\eta$ in (\ref{eq1}) may abruptly change through an {\em active
contact}; the notion of active contact is illustrated in the simple case of a neuron receiving a single spike in Appendix \ref{illustration}. The potential $V_i^*(t)$ may also touch the threshold without crossing it, and the noise may remain constant over some time interval; we call such an event {\em passive contact}. That the potential can brush, or remain at the threshold level without producing a spike is made possible by the $\sigma\to 0$ limit.
We will discuss later on the validity of this calculation, and how to modify it
when the noise standard deviation, $\sigma$, does not vanish.
Both types of contacts are shown in Fig.~\ref{fig-scheme}A.


Let us explain how the positions of active and passive contacts can be
determined.  Let $t_1<t_2< \ldots <t_M$ be the emission times of the
spikes arriving from the neurons interacting with $i$ during the time
interval $[t_0 \equiv t_{i,k};t_{M+1} \equiv t_{i,k+1}]$, and
$J_1,J_2,\ldots , J_M$ the corresponding synaptic
strengths\footnote{Due to the limited temporal resolution of the
measurement two inputs of amplitudes $J$ and $J'$ can apparently
arrive at the same time; if so, we consider, based on model
(\ref{ode}) and (\ref{defisyn}), that a single input of amplitude $J+J'$ enters the
neuron.}. Let $V_0=0$ be the initial value of the potential, and $m_0=1$ be the index of the first input spike. If the time is small enough the optimal potential is surely below the threshold value. According to (\ref{eq1}) the optimal noise is an exponential with noise coefficient $\eta$, and the optimal potential is obtained by solving (\ref{optimb}) with the result,
\begin{eqnarray}\label{defv}
V_i(\eta, t) &=& V_0 \;e^{-(t-t_0)/\tau} +
\sum _{m=m_0}^M \frac{J_m}C\; e^{-(t-t_m)/\tau} \theta(t-t_m) \nonumber \\
&+& \frac{I_i}g (1-e^{-(t-t_0)/\tau}) 
+ \frac{\eta}{g}\, \sinh\left( \frac{t-t_0}\tau\right)
\end{eqnarray}
where $\theta$ is the Heaviside function. It is tempting to look for the value of $\eta$ such that a spike is emitted at time $t_{M+1}$, defined by the implicit equation $V_i(\eta,t_{M+1})=V_{th}$. However, the corresponding potential might not be below threshold at all intermediate times $t_0<t<t_{M+1}$. Instead, we look for the smallest noise capable of driving the
potential from its initial value $V_0$ into contact with
the threshold:
\begin{equation}\label{defnoise}
\eta^* = \min \big\{ \eta : \max_{t_{0}<t\le t_{M+1}}
V_i(\eta ,t) =V_{th} \big\} \ .
\end{equation}
As the potential (\ref{defv}) is a monotonically increasing function of $\eta$, a value of the noise smaller than $\eta^*$ would not be able to bring the potential to the threshold and to trigger a spike at any time, while a value larger than $\eta^*$ would violate the condition that the potential cannot cross the threshold on the time interval $t_0<t<t_{M+1}$, see Appendices \ref{seccp-sec} and \ref{seccp2}.

We denote by $t_c$ the time at which the threshold is reached: $V_i(\eta^*,t_c)=V_{th}$.  
The solution to the minimization problem (\ref{defnoise}) can be found following the procedure described below. Briefly speaking, the procedure identifies candidates for the contact points, selects the best one, and is iterated until the ISI is completed.
\begin{itemize}
\item {\em Active candidates:} we first consider the possibility that the contact time $t_{c}$ coincides with a synaptic input. We therefore calculate for each $m=m_{0},\ldots, M+1$, the root $\eta_{m}$ of the  implicit equation $V(\eta_{m},t_{m})=V_{th}$. The smallest of those $M$ noise coefficients is called $\eta ^*_{a}$. 
\item{\em Passive candidates:} we then consider the case where the contact time $t_{c}$ may not be simultaneous to any input, but rather fall between two successive spikes. For each $0\le m\le M$, we look for a  noise coefficient $\eta_{p}$ and a contact time $t_{c}\in [t_{m};t_{m+1}]$ fulfilling the set of coupled equations $V_i(\eta_{p},t_c)=V_{th}, \dot V_i(\eta_{p},t_c)=0$, expressing that the potential reaches and does not cross the threshold. These two equations can be solved analytically, see expressions (\ref{etap}) and (\ref{etap2}) in Appendix \ref{secgnon}. We call $\eta_{p}^*$ the smallest noise coefficient corresponding to those possible passive contacts.
\item{\em  Selection of the best candidate:} 
\subitem{$\bullet$} if $\eta^*_{a} < \eta^*_{p}$, the contact is active and takes place at time $t_{c} =t_{m^*}$ for a certain $m^*$ comprised between $m_0$ and $M+1$ (Fig.~\ref{fig-scheme}A). The optimal potential and noise in the time interval $[t_0,t_{m^*}]$ are given by, respectively, eqns (\ref{defv}) and (\ref{eq1}) with $\eta=\eta^*$. 
\subitem{$\bullet$} If $\eta^*_{p} < \eta^*_{a}$, the contact is passive, and takes place in the time interval $[t_{m_{c}-1};t_{m_{c}}]$ for a certain $m_{c}$ comprised between $m_0$ and $M+1$ . The potential will remain equal the threshold, and the noise will remain constant according to (\ref{eq2}) over a finite time interval $[t_{c};t_{c}+\Delta_c]$, after which both $V_i^*(t)$ and $\eta_i^*(t)$ resume their course
(Fig.~\ref{fig-scheme}A).  $\Delta_{c}$ is the smallest delay allowing
the potential to be in active contact with the threshold at a later
time $t_{m^*}$, with $m^*\ge m_c$. The correctness of this statement is ensured by the fact that there can be at most one passive contact
between two active contacts (Appendix \ref{secgnon}); hence, a passive contact is necessarily followed by an active contact (or by the spike at the end of the ISI). For every integer $m$ comprised between $m_c$
and $M+1$, we calculate analytically the delay $\Delta_c(m)$ such that the potential reaches the threshold in $t_{m}$, see eqn (\ref{eqdeltac}) in Appendix \ref{secgnon}; the smallest among those delays and the corresponding value of $m$ are,
respectively, $\Delta_c$ and $m^*$.  
\item{\em Iteration:} we are left with the calculation of $\eta_i^*(t)$ and $V_i^*(t)$ on the remaining part of the inter-spike interval, $[t_{m^*};t_{M+1}]$. To do so, we iterate the previous steps. We first update $t_0\leftarrow t_{m^*}$, $m_0 \leftarrow m^*+1$,
$V_0 \leftarrow V_{th}+\theta(-J_{m^*})\frac{J_{m^*}}C$ in (\ref{defv}), and
look for the lowest noise producing a new contact over
the interval $[t_0, t_{M+1}]$ using (\ref{defnoise}) again. The procedure
is repeated until the whole inter-spike time interval is exhausted. 
\end{itemize}
As a result a sequence of values for $\eta^*$ is built, 
each value corresponding to the noise coefficient (\ref{defnoise}) 
between two successive active contact points.

\subsection{How small should the variance of the noise be?}

The LIF dynamical equation (\ref{ode}) involves quantities, such as the membrane
potential, the membrane conductance, the input current, which have different
physical units. A straightforward algebra shows that (\ref{ode}) is equivalent
to the following differential equation,
\begin{equation}
\frac{d\bar V}{d\bar t}= -\bar V + \sum_{j(\ne i)} \bar J_{ij} \sum _k \delta 
(\bar t -\bar t_{j,k}) +\bar I_i + \bar \eta _i (\bar t) \ ,
\end{equation}
which involves only dimensionless variables (denoted with overbars):
$\bar t = t\;\frac gC ,\ \bar V = \frac{V}{V_{th}},\ 
\bar J_{ij} = \frac{J_{ij}}{C\, V_{th}},\ 
\bar I_i = \frac{I_i}{g\, V_{th}}.$ The noise has zero mean, and
covariance $\langle \bar \eta_i(\bar t)\bar \eta _j (
\bar t') \rangle = \bar \sigma^2 \delta _{ij}\delta (\bar t-\bar t')$,
where 
\begin{equation}\label{defnounit}
\bar \sigma = \frac{\sigma}{V_{th}\sqrt{gC}} \ .
\end{equation}
Intuitively, we expect that the potential $V_i(t)$ will not depart much from the
optimal path $V_i^*(t)$, and, hence, that our inference algorithm will
be accurate if the dimensionless standard deviation of the noise, $\bar \sigma$, is small. 
We illustrate this claim on the simple case of a neuron receiving a few
inputs from two other neurons during two inter-spike intervals (ISI)
of length $2\tau$, see Fig.~\ref{fig-compar-optpath}B. The times
of the input spikes were randomly chosen, once for all, before the simulations
started. Then, we numerically integrated the LIF equation for the
potential (\ref{ode}) for $10^6$ random realizations of the noise $\eta(t)$. The realizations such that the neuron spiked twice, with ISIs falling in the range
$[1.99,2.01]\times \tau$ were considered as successful. The number
of successful realizations was comprised between $10^3$ and $10^4$,
depending on the noise level, $\sigma$. We show in 
Fig.~\ref{fig-compar-optpath}B the paths of the potential and of the noise,
averaged over successful realizations, and compare them
to the optimal potential, $V^*$, and  noise, $\eta ^*$.
As expected the agreement is very good for small $\sigma$. 
We now make this observation quantitative.

Consider the $k^{th}$ inter-spike interval $[t_{i,k};t_{i,k+1}]$ of neuron $i$. 
The optimal potential $V_i^*(t)$ is the time-course followed by
the LIF membrane potential $V_i(t)$ in the $\sigma \to 0$ limit. When the noise
variance is not vanishing, the potential $V_i(t)$ can slightly 
deviate from the optimal path (Fig.~\ref{fig-fluctuphi}C).
Deviations are null at the extremities of the inter-spike interval due
to the boundary constraints on the potential. A measure of the
magnitude of the fluctuations of the potential is thus given by
the variance of $V_i(t)-V_i^*(t)$ at the middle of the ISI, {\em i.e.}
$t=\frac 12(t_{i,k}+t_{i,k+1})$  (Fig.~\ref{fig-fluctuphi}C). This variance can be calculated when the constraint that the fluctuating potential $V_i (t)$ does not cross the threshold at times $t<t_{i,k+1}$ is relaxed, see  Appendix \ref{app-corrections}. We obtain 
\begin{equation} \label{result-fluctu}
\frac{\langle (V_i -V_i^*)^2\rangle}{V_{th} ^2} = \bar \sigma ^2 \;
\tanh\left( \frac{t_{i,k+1}-t_{i,k}}{2\tau} \right) ,
\end{equation} 
where $\tau$ is the membrane leaking time. As expected, if $\bar\sigma$ is 
small, so are the fluctuations of the potential around the optimal path. 

However, the reverse statement is false. Consider, for instance, 
the case of a perfect integrator, for which the dimensionless $\bar\sigma$
(\ref{defnounit}) is virtually infinite. 
Sending $g\to 0$ in (\ref{result-fluctu}),  we obtain 
\begin{equation} \label{result-fluctu2}
\frac{\langle (V_i -V_i^*)^2\rangle}{V_{th} ^2} = \frac{ \sigma^2\,
\,(t_{i,k+1}-t_{i,k}) }{2 \,(C\, V_{th})^2} \qquad \qquad (g\to 0)\ .
\end{equation} 
Hence, the relative fluctuations of the potential are small if the
typical amplitude of the electrical charge entering the neuron during 
the ISI due to the noise, $\sigma
\sqrt{t_{i,k+1}-t_{i,k}}$, is small compared to the total charge $C\, V_{th}$ 
necessary to reach the threshold from the rest state. 
It is interesting to note that this statement applies to the LIF, too.
Whatever the level of the noise, $\bar \sigma$,  the relative fluctuations of the potential (\ref{result-fluctu}) can be made small if the  duration of the ISI is short enough compared to the membrane leaking time, $\tau$.

\subsection{Beyond the weak-noise limit: the Moving Threshold procedure}
\label{sec-beyond}

For large values of $\sigma$, a discrepancy between the optimal potential and the potential obtained in simulations appears (Fig.~\ref{fig-pro}B). A general observation is that the optimal potential calculated by the Fixed Threshold procedure can get very close to $V_{th}$, while the true potential stays further away from the threshold to avoid premature firing. To further illustrate this effect, consider a system of two IF neurons, 1 and 2, both fed with an external current. In addition, neuron 1 receives positive inputs from neuron 2 ($J_{12}>0$), and neuron 2 is independent from the activity of neuron 1 ($J_{21}=0$).  In presence of a strong noise, the optimal potential calculated from the Fixed Threshold procedure quickly reaches a stationary value close to $V_{th}$, while the random potential obtained from simulations fluctuates around a much lower level (Fig.~\ref{fig-pro}E). The presence of a strong noise biases the membrane potential to lower values to prevent early spiking. A heuristic approach to reproduce this bias consists in decreasing the threshold from $V_{th}$ to a time- and context-dependent value, $V_{th}^M$. We now explain how this moving threshold, $V_{th}^M$, is determined.

Consider first a neuron with no synaptic input, fed with an external current $I$, during the inter-spike interval $[t_{i,k};t_{i,k+1}]$.  We call $p_s(\delta t|V)$ the probability that the potential, taking value $V$ at time $t_{i,k+1}-\delta t$, remains below the threshold at any larger time $t$, with $t_{i,k+1}- \delta t<t< t_{i,k+1}$. This probability depends on the current $I$, and can be expressed for an arbitrary level of noise, $\sigma$, as a series of parabolic cylinder functions (Alili, Patie and Pedersen, 2005).  Figure \ref{fig-pro}D show $p_s$ as a function of $V$ for some characteristic values of the parameters. The probability of survival, $p_s$, sharply decreases to zero when $V$ gets close to the threshold, $V=V_{th}$. We model this outcome by the following approximation, which involves a new, effective threshold $V_{th}^M$: we consider that the processes starting from a value of the potential $V>V_{th}^M$ will not survive for a time delay $\delta t$.  In other words, the true threshold, $V_{th}$, is changed into a 'moving' threshold, which is a function of the current $I$, the time $\delta t$, and the parameters $g,C,\sigma$.  A simple way to
define $V_{th}^M$ is to look at the intersection of the tangent line to $p_s$ in $V=V_{th}$ with, say, the $p_s=\frac 12$ line\footnote{This choice is arbitrary; other values, ranging from $\frac 14$ to $1$ have been tried, do not qualitatively affect the results presented later in this article.}; the resulting expression for $V_{th}^M$ is given in Appendix \ref{essai2}. Figure \ref{fig-pro}E shows the output of the Moving Threshold procedure on the simple 2-neuron system described above. The optimal potential, 'pushed' down by the moving threshold $V_{th}^M$ is much lower than in the Fixed Threshold approach and in much better agreement with the random realization of the membrane potential. More details are given in Section \ref{secsubmt}.  

To account for the existence of synaptic inputs, we may choose the parameter $I$ entering the calculation of $p_s$ and $V^M_{th}$ to be the value of the effective current $\displaystyle{I_i^e = I_i + \sum _{j(\ne i)} J_{ij} \, f_j}$, rather than the external current $I_i$ itself. Here, $f_j$ is the average firing rate, defined as the number of spikes fired by neuron $j$ divided by the duration $T$. Contrary to the external current $I_i$, the effective current $I_i^e$ takes into account the (average) input current coming from other neurons.  This choice was done in the numerical experiments reported in the Results section. To further speed up the calculations, we derive the value of $V_{th}^M$ for discrete-values delays $\delta t$ only; in a discrete interval, $V_{th}^M$ is kept to a constant value.

Alternative heuristic approaches to deal with the presence of moderate noise can be proposed. In Appendix \ref{essai2} we introduce a cost-function for the effective current, whose effect is also to decrease the optimal potential. These approaches are effective when the optimal potential calculated by the Fixed Threshold procedure quickly saturates to a level close to $V_{th}$. More precisely, we expect the Moving Threshold procedure to be efficient if the membrane leaking time is smaller or comparable to the ISI, and the leaking current, $\simeq gV_{th}$, is larger or equal to the external current, $I$.

\subsection{Maximization of the log-likelihood to infer the interactions and currents}

The Fixed or Moving Threshold procedures allow us to calculate the optimal paths for the potential and the noise, given the couplings and currents. Knowledge of those paths gives us also access to the 
logarithm of the likelihood $P$ in the $\sigma \to 0$ limit, 
\begin{eqnarray}\label{deflpcc}
 L^* ({\cal T}|{\cal J},{\cal I}) &=&
\lim _{\sigma \to 0} \ \sigma^2 \; \log P ({\cal T}|{\cal J},{\cal I})\nonumber \\
&=&\sum_{i,k} {\cal L}[ V_i^*(t); k,{\cal T},{\cal J},{\cal I}]
=-\frac 12 \sum _{i,k} \int _{t_{i,k}} ^{t_{i,k+1}} dt \, \eta_i^*(t)^2
\end{eqnarray}
Since $L^*$ in (\ref{deflpcc}) involves the sum over different neurons,
the maximization over the couplings $J_{ij}$ and the current $I_i$ of neuron
$i$ can be done independently of the other couplings $J_{i'j}$ and 
currents $I_{i'}$ ($i'\ne i$). Formally, we are left with $N$ independent
inferences of the most likely couplings and current for a single
neuron, in presence of the spikes emitted by the $N-1$ other neurons. 
As a consequence neurons `decouple' in the inverse problem: 
the couplings $J_{ij}$ and the current $I_i$ of neuron
$i$ can be inferred independently of the other couplings $J_{i'j}$ and 
currents $I_{i'}$ ($i'\ne i$).

${\cal L}$ defined in (\ref{pathint2}) is a negative-semidefinite 
quadratic function of its arguments $V_i(t), J_{ij},I_i$. It is thus a
concave function of the  couplings and the currents. This property holds for  $L^*$ (\ref{deflpcc}) (Boyd and Vandenberghe, 2004). In order to infer the most likely current $I_i$ and couplings $J_{ij}$, we start
from an arbitrary initial value {\em e.g.} $I_i=J_{ij}=0$. The full path of
the optimal noise, $\eta^*_i(t)$, over all the inter-spike intervals
$k$ of neuron $i$, is calculated following the above procedure. 
We then update the couplings and the current using the Newton-Raphson
method to maximize $\log P$, {\em i.e.} to minimize  
the integral of the squared optimal noise, see (\ref{deflpcc}). 
Convergence follows from the concavity property stated above.
The procedure requires the expressions for the 
gradient and the Hessian matrix of $\log P$ with respect to 
the couplings $J_{ij}$ and the current $I_i$, which can be calculated 
exactly from (\ref{deflpcc}) and (\ref{defv}). Note that $\log P$ is piecewise continuously twice-differentiable; while the gradient is continuous for all $J_{ij}$ and $I_i$,  the Hessian matrix is bounded and negative, and may discontinuously jump due to a change of the contact points. Knowledge of the
Hessian matrix is also important to determine how reliable are the
values of the inferred parameters.

\subsection{Accuracy on the inferred parameters}\label{sec-accuracy}

When the variance of the noise, $\sigma ^2$, 
vanishes the inferred parameters cannot deviate from
their most likely values. However, for small but non zero $\sigma$, 
deviations are possible\footnote{Note that the inferred parameters might be less
sensitive than the time course of the potential to the noise level $\sigma$.
The reason is that the corrections to the log-likelihood $L^*$, to the lowest order
in the noise variance $\sigma^2$, do not depend on the current and 
interactions (Appendix \ref{app-corrections}).}. The probability for such
deviations can be estimated from the expansion 
of $L^*$ around its maximum. We introduce for each
neuron $i$, the $N$-dimensional vector $v_i$ whose components
are: $v_i^{(i)} = I_i\;\tau$, and $v_{j}^{(i)}= J_{ij}$ for $j \ne
i$. The multiplication of the current by the membrane leaking time
ensures that all components can be expressed in units of a coupling.
Similarly we call $\hat v^{(i)}$ the vector obtained when the current and
couplings take their most likely values, that is, maximize $L^*$. 
Let us 
call
\begin{equation} \label{fluch}
{\bf H} ^{(i)} _{j,j'} = -\frac 1{\sigma^2} \frac{\partial ^2 L^* }{\partial
  v_j ^{(i)} \partial v_{j'} ^{(i)}}({\cal T}| \hat {\cal J},\hat {\cal I}) \ .
\end{equation}
the Hessian matrix of $L^*$. The parameters $v_j^{(i)}$ are normally
distributed around their most likely values, with a covariance matrix
given by
\begin{equation} \label{flucov}
\langle \big(v^{(i)}_j - \hat v^{(i)}_j\big) 
 \big(v^{(i)}_{j'} - \hat v^{(i)}_{j'}\big) \rangle = 
\big[ {\bf H}  ^{(i)}\big] ^{-1}_{j,j'} \ .
\end{equation}
In particular, the error bars on the inferred parameters are given by the 
diagonal elements of the inverse of ${\bf H}^{(i)} $. Note that, if
the value of $\sigma$ is not known, formulas (\ref{fluch}) and (\ref{flucov}) 
can still be used to compare the error bars between each other.

As the entries of ${\bf H}^{(i)} $ scale linearly with the duration
$T$ of the recording, or, more precisely, the number $S$ of recorded
spikes the uncertainty on the inferred parameters will decrease as
$S^{-1/2}$. 
A detailed spectral analysis of $\sigma^2\,{\bf H}^{(i)} /S$ in the case of weak couplings, reported in Appendix
\ref{aeigen}, shows that the largest eigenvalue, $\lambda _{max}$,  is related to the
fluctuations of the effective current, 
\begin{equation}\label{defieff}
I^e_i = I_i +\sum _{j (\ne i)} J_{ij}\, f_j^{\,i,\tau}  \ ,
\end{equation} 
where
\begin{equation}\label{deffjitau}
f_j ^{\, i,\tau} = \frac 1T \sum _{k,\ell:t_{i,k}<t_{j,\ell}<t_{i,k+1}} \exp 
\left( -\frac{t_{i,k+1}-t_{j,\ell}}\tau\right)
\end{equation}
is the average firing rate of neuron $j$,
calculated over the time scale $\sim\min(\tau,ISI)$ preceding a spike
of neuron $i$. The smallest eigenvalue, $\lambda_{min}$, 
corresponds to the fluctuations of the current $I_i$ alone. In other words, the
uncertainty on the inferred value for $I_i^e$ is much smaller than the one
on the current $I_i$. The intermediate eigenmodes describe correlated 
fluctuations of the couplings. 
Explicit expressions for the largest and smallest eigenvalues, 
$\lambda_{max}$ and $\lambda_{min}$, are derived in Appendix~\ref{aeigen}.

When  a
small change of ${\cal J}$ and ${\cal I}$ causes a modification of the set of contact points the second derivative of $L^*$ may be discontinuous. A simple illustration is provided by the the case of a single input, whose
 log-likelihood $L^*$ is reported in Appendix \ref{illustration}. 
If the maximum is located at, or very close to the boundary dividing two or more
sets of contacts, the value of the Hessian matrix will depend on the direction 
along which the maximum $\hat {\cal J},\hat {\cal I}$ is approached. 
This phenomenon is also encountered in the analysis of real data, see Section \ref{sec-inf-neg}.

\section{Results}\label{sec-test}

\subsection{Tests on simulated data}

In this Section, we test our inference procedure on synthetic data
generated from networks with known interactions and currents. 
We compare the results obtained from our two inference algorithms,
the Fixed and Moving Threshold procedures, respectively defined in Sections
\ref{sec-algo} and \ref{sec-beyond}.

\subsubsection{Scaling of the computational time}

\begin{figure}[t]
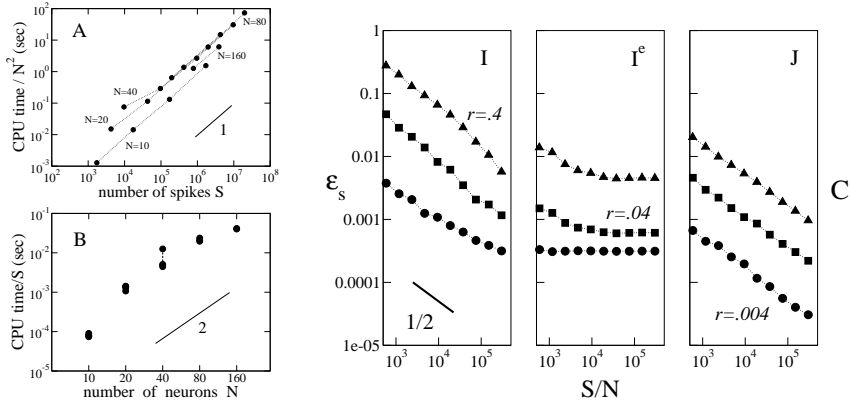

\begin{center}
\epsfig{file=fig3ab.eps,width=3.6cm} 
\hskip .5cm
\epsfig{file=fig3c.eps,width=6.9cm}
\caption{Results of the Fixed Threshold algorithm on
a network of $N$ uncoupled neurons, and in the absence of leakage.
The running time, on one core of a 2.8 GHz Intel Core 2 Quad desktop
computer, is shown as a function of the number of spikes, $S$ ({\bf A}), 
and of the number of neurons, $N$ ({\bf B}).
{\bf C.} Inference errors $\epsilon _s$ on the currents $I_i$,
$I^e_i$, and on the couplings $J_{ij}$ vs. $S/N$, for $N=40$ neurons and 
three values of the noise ratio (\ref{defrici}):
$r=.4$ ($\blacktriangle$), .04 ($\blacksquare$), .004 ($\bullet$). Data
are shown for one randomly drawn sample; sample-to-sample fluctuations
are of the order of the symbol size.  
Full lines show square root, linear and quadratic increases (in log-log scale);
dotted lines serve as guides to the eye.}
\label{fig-time}
\label{fig-error}
\end{center}
\end{figure}

We first consider $N$ (ranging from 20 to 160) neurons, with no
leakage ($g=0$). The neurons are uncoupled ($J_{ij}=0$ for $i\ne j$),
and fed with identical currents ($I_i=I$ for all $i$).
The choice of the noise variance, $\sigma^2$, is specified later. 
The LIF equation is solved numerically, using a fourth-order 
Runge-Kutta integration scheme. We choose the elementary time step 
to be $10^{-5}$ sec, while the average duration of the ISI is
$10^3$ to $10^5$ longer. For each realization of the noise, the simulation
is run until a set of $\simeq 10^7$ spikes is generated. We then use
the first $S$ spikes in this set to infer the currents and the couplings 
(not fixed to zero {\em a priori}) with the Fixed Threshold procedure. 
The algorithm stops if the log-likelihood $L^*$ increases by less than $\epsilon =10^{-12}$
after an iteration of the Newton-Raphson procedure. Alternatively, 
the algorithm may halt when the overall change in the couplings 
and current becomes smaller than a certain {\em a priori} bound.

Figures \ref{fig-time}A\&B show how the running time scales with, respectively, 
the number $S$ of spikes, and the number $N$ of neurons. The empirically found
scaling, $O(S\; N^2)$, can be understood as follows.
Consider one neuron, say, $i$. The number of spikes of neuron $i$ is, on 
average, equal to $S/N \simeq f \;T$, where $T$ is the duration of the recording and
$f$ is the average firing rate. The number of contact points, $N_{co}$, 
is found to scale as the number of spikes, $S/N$. 
The calculation of the contribution to the Hessian 
${\bf H}^{(i)}$ coming from the interval between two successive contact points 
of $V^*_i$ takes $O(N^2)$ time. The total calculation of 
${\bf H}^{(i)}$ thus requires $N_{co}\; N^2\simeq S\;N$ 
operations\footnote{Note that the 
ratio of the time to calculate ${\bf H}^{(i)}$ over the time required for 
the inversion of the Hessian matrix is equal to $N_{co}\; N^2/N^3
\sim S/N^2$, and is generally much larger than one. The reason is that the 
number of parameters to be inferred, $N$, has to be smaller than the number
of constraints over the optimal potential, $N_{co}$. For the 
real data analyzed in Section \ref{sec-real}, we have $S/N^2\simeq 64 $ and
$108$ for, respectively, Dark and Natural Movie data sets.}.  
The loop over the neuron index, $i$, gives an extra (multiplicative) factor $N$. 

The running time of the Moving Threshold algorithm grows as $S\; N^2$, 
too. However the proportionality constant is generally
higher than for the Fixed Threshold procedure, due to the extra 
computational burden to calculate $V_{th}^M$. For fixed $N$ and $S$, the
running times of both procedures increase with the number of contacts, 
{\em e.g.} when the membrane conductance $g$ increases. This effect 
is described in Section \ref{sec-real}.

\subsubsection{Dependence of the inference error on the number of spikes}

We define the inference errors as the root mean square of the difference 
between the inferred parameters, $J_{ij}^{inf}, I_i^{inf}$ and the true values,
$J_{ij}=0,I_i=I$: 
\begin{equation} 
\epsilon _s (J) = \sqrt{\frac 2{N(N-1)} \sum _{i<j} 
\left(\frac{J_{ij}^{inf}-J_{ij}}{CV_{th}}\right)^2}
\ ,\quad \epsilon _s (I) = \sqrt{\frac 1{N} \sum _{i} 
\left(\frac{I_{i}^{inf}}{I}-1\right)^2}
\  ,
\end{equation}
together with a similar definition for the effective current, $\epsilon(I_i^e)$,
with $I_i^{inf}$ replaced with the inferred value for $I^e_i$.
The inference errors depend on the dimensionless noise ratio\footnote{When 
$g=0$, changing the value of the current $I$ amounts to changing the time-scale of the evolution of the potential in (\ref{ode}). Hence, the errors $\epsilon_s$ depend on the parameters $I,C,\sigma,V_{th}$ through the value of $r$ only (as long as $I>0$).},
\begin{equation}\label{defrici}
r = \frac{\sigma}{\sqrt{I\, C\, V_{th}}} \ . 
\end{equation} 
Figure \ref{fig-error}C shows the
inference errors found for different noise ratios $r$, and their dependence
on the number $S$ of spikes, in the absence of membrane leakage.
For small data sets, the inference error is mainly due to the imperfect sampling.
As the number $S$ of spikes increases, $\epsilon _s$ decreases as $S^{-1/2}$, 
as expected from Section \ref{sec-accuracy}. 
When $S$ is very large, the errors saturate to a residual value, $\epsilon_\infty$.  The
presence of the residual error $\epsilon _\infty$ results
from the dominant-path approximation done in our calculation of the likelihood $P$. 
The value of $\epsilon_\infty$ decreases with $r$ as expected. 

The cross-over between the sampling-dominated and residual error regimes
takes place for a certain value of the number of spikes, $S_{c.o.}$. 
Both $S_{c.o.}$ and $\epsilon_\infty$ depend on the observable, {\em i.e.}
$I,I^e,J$, and on the noise ratio $r$. With the values of $S$ reached in the 
simulations, the onset of the cross-over is clearly visible for $I^e$, 
can be guessed for $I$, and is not observable for $J$.
The existence of a cross-over, and an estimate of $S_{c.o.}$ 
can be derived from the discussion of Section \ref{sec-accuracy}. When
$S$ is large, the {\em a posteriori}
distribution of the inferred parameter, $v=I,I^e,$ or $J$, becomes Gaussian, 
with a variance
\begin{equation}
\langle (\Delta v)^2 \rangle \simeq \frac{\sigma^2}{\lambda\; S} \ , 
\end{equation}
where $\lambda$ is the eigenvalue of the Hessian matrix of $L^*$ attached to
the fluctuations of the parameter $v$.
The inference error sums up contributions coming from both the sampling
fluctuations and the residual error. The cross-over takes place when both 
contributions are comparable, $\langle (\Delta v)^2 \rangle^{\frac 12} =
\epsilon _\infty$, that is, for
\begin{equation}
S_{c.o.}\sim\frac{\sigma^2}{\lambda\;\epsilon_\infty^2} \ . 
\end{equation}
Figure~\ref{fig-error}C confirms that $S_{c.o.}$ diminishes with
$\sigma$ (or $r$), and is much smaller for $I^e$ than for $I$ (as expected from
the dependence on the eigenvalue $\lambda$);  moreover, the
residual error on the couplings is extremely small (or might be even zero).

As a conclusion, our inference algorithm is very accurate in the absence of membrane
leakage. With $10^3$ spikes per neuron only and $r=.004$, for instance, the errors on the
currents and on the couplings are, respectively, $\epsilon_s=3 \; 10^{-3}$ and 
$4\; 10^{-4}$. Even in the presence of strong noise ($r=.4$), and with the same number of
spikes per neuron, the errors on the effective currents and on the couplings are less than 1\%.

\subsubsection{Performance of the Fixed Threshold procedure on networks of coupled neurons}

\begin{figure}[t]
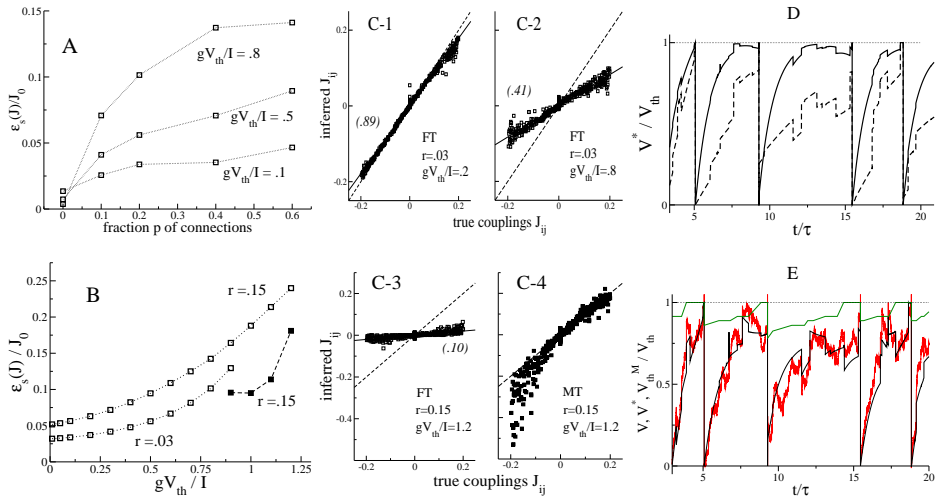

\begin{center}
\epsfig{file=fig4a.eps,width=4cm}
\epsfig{file=fig4c12.eps,height=3cm,width=4cm} 
\epsfig{file=fig4d.eps,width=4cm}
\vskip .3cm
\epsfig{file=fig4b.eps,width=4cm}
\epsfig{file=fig4c34.eps,width=4cm}
\epsfig{file=fig4e.eps,width=4cm}
\caption{Results from the Fixed (empty squares) and Moving (full squares) 
Threshold algorithms on a random network of $N=40$ coupled neurons;
the maximal amplitude of synapses is $J_0=.2\, CV_{th}$. 
The error on the couplings, $\epsilon _s(J)/J_0$,
is plotted as a function of the fraction $p$ of connections ({\bf A}) and
conductance over current ratio, $gV_{th}/I$ ({\bf B}). 
Each simulated data set contains $S=5\; 10^5$ spikes, which is larger
than the cross-over size $S_{c.o.}$; the symbol size correspond to the fluctuations
estimated from ten different data sets for the same network of interactions.
{\bf C.}  Inferred interactions vs. true values of $J_{ij}$ for various
values of $gV_{th}/I$ and $r$, and for one random network with a fraction $p=.2$
of connections.  Dashed lines have slope unity.  Panels C-1, C-2, C-3
show the results of the Fixed Threshold (FT) procedure; the slopes of
the best linear fits (full lines) are indicated between
parenthesis. Panel C-4 shows the outcome of the Moving Threshold (MT)
procedure; even if multiplied by $10$, the FT couplings of Panel C-3 are in much
worse agreement with the true interactions than the MT couplings.
{\bf D.}~Optimal potentials $V^*$ obtained with the 
Fixed Threshold procedure for $g=.1 \, I/V_{th}$ (dashed curve) and 
$g=1.2 \, I/V_{th}$ (full curve), and for one arbitrarily chosen neuron among the 
$N=40$ neural cells; the noise ratio is $r=.15$. 
{\bf E.} Comparison of a random realization of the potential $V$ (red)
with the optimal potential $V^*$ (black) obtained with the
Moving Threshold $V_{th}^M$ (green) procedure.
The network of interactions, the spiking times, and the arbitrarily chosen neuron are the
same as the ones in {\bf D} for $g=1.2 I/V_{th}$. The time-average of $V_{th}^M$ is
$\simeq .93 \; V_{th}$.}
\label{fig-errorvsp}
\label{fig-error-inf-true}
\label{fig-coupled}
\label{fig-mb0}
\label{fig-mb}
\label{fig-negpos}
\end{center}
\end{figure}

We now study the ability of the algorithm to infer the interactions between 
coupled neurons. To do so, we consider random connection graphs built in the
following way (Bollobas, 2001). 
We start from a complete oriented graph over $N$ neurons,
and erase each one of the $N(N-1)$ link with probability $1-p$, 
independently of each other. The removal process is not symmetric:
the link $i\to j$ may be removed, while the connection $j\to i$ is preserved. 
At the end of the construction process, the average number of outgoing (or incoming) 
neighbors of a neuron is $p(N-1)$. Each existing connection is then assigned
a synaptic weight, uniformly at random  over the interval $[-J_0;J_0]$. 
All neurons receive the same external current $I$. In addition, the
membrane conductance, $g$, is now different from zero.
The values of $p, J_0, I,g$, and $\sigma$ are chosen so that the network remains below
saturation.

We have also performed simulations where the interaction graph is drawn
as above, but each neuron $i$ is chosen to be either excitatory or
inhibitory with equal probabilities. The outgoing interactions from
$i$ have all the same sign, and random amplitudes in $[0;J_0]$. The
performance of our inference algorithms are qualitatively similar for both models.

Figure \ref{fig-errorvsp}A shows the error on the couplings inferred with the
Fixed Threshold algorithm,
$\epsilon _s(J)$, as a function of the fraction $p$ of connections, for three values
of the membrane conductance over current ratio. The error roughly increases
as $\sqrt{p}$, that is, the number of connections in the network. This scaling 
suggests that much of the inference error is due to non-zero couplings. 
This finding agrees with Fig.~\ref{fig-error}C, 
which showed that the inferred interactions between uncoupled neurons was very small
in the $g=0$ case. To better understand the performance of the algorithm, we compare in 
Fig.~\ref{fig-error-inf-true}C the inferred interactions $J_{ij}$ with their 
true values for the 1560 oriented pairs $j\to i$ of a randomly drawn network of 
$N=40$ neurons, with $p=.2$ and $J_0=.2 CV_{th}$. When the ratio $gV_{th}/I$ is small
compared to unity, the quality of the inference is very good 
(Fig.~\ref{fig-error-inf-true}C-1). For larger ratios  $gV_{th}/I$  the
inferred couplings are still strongly correlated with their true values, but 
are approximately rescaled by an overall factor $<1$, corresponding to the 
average slope of the linear regression in Fig.~\ref{fig-error-inf-true}C-2. 
As $gV_{th}/I$  increases, this factor decreases and the inference error
grows (Fig.~\ref{fig-errorvsp}A).

Figure~\ref{fig-coupled}B shows that the inference error on the interactions 
increases not only with $gV_{th}/I$ but also with the noise ratio $r$.
For large values of $r$, the network can sustain activity even when $g V_{th}>I$,
and the inference error can take large values (upper curve in 
Fig. \ref{fig-coupled}B). In this regime, the couplings found by the
Fixed Threshold algorithm become small, and the inferred current $I_i$ gets close
to $gV_{th}$. The corresponding potential $V^*_i(t)$ rises sharply, in a time
$\tau$, to a value slightly below threshold, $I_i/g$, with small
fluctuations due to the synaptic inputs. This phenomenon can be seen in
Fig.~\ref{fig-mb0}D, which compares the optimal potential of
a neuron for two different values of membrane conductance.
As discussed in the Methods section, this behavior is a consequence of
the $\sigma\to 0$ limit taken in the calculation of the optimal potential; 
when $\sigma$, or $r$, is not small, the potential is
unlikely to stay close to the threshold for a long time without
producing a spike, see Fig.~\ref{fig-compar-optpath}B. In the next 
paragraph, we analyze the results of the Moving Threshold inference procedure.

As a conclusion, zero couplings are perfectly inferred, while the amplitude
of large (positive or negative) interactions can be underestimated by the
Fixed Threshold algorithm, especially so when the noise is strong. 
However, the relative ordering of the interactions is essentially 
preserved by the inference procedure.

\subsubsection{Inference error with the Moving Threshold procedure}
\label{secsubmt}

The Moving Threshold procedure was tested in Fig.~\ref{fig-pro}E on an asymmetric system of two IF neurons ($J_{12}/( CV_{th})=.1,J_{21}=0$) in the presence of a strong noise, see description in caption and Section \ref{sec-beyond}. While the Fixed Threshold procedure erroneously inferred that both interactions vanish, the Moving Threshold correctly inferred the sign and the order of the magnitude of the coupling: $J_{12}^{inferred}/(CV_{th})=.2 \pm .1$. The inferred currents were within 10\% of their true values. These results were obtained from a large number $S$ of spikes to avoid finite-$S$ effects.

The synthetic data used in Fig.~\ref{fig-coupled}B were generated with two different values of the noise ratio, $r$. We estimate the relative fluctuations of the potential
around the optimal path, averaged over all the inter-spike intervals in the data set, using formula (\ref{result-fluctu}), and find
\begin{equation}\label{result-fluctu-num}
\frac{\sqrt{\langle (V_i -V_i^*)^2\rangle}}{V_{th} } \simeq \left\{
\begin{array} {c c c}.028 & \hbox{\rm for}& r=.03\\ .138& \hbox{\rm for}& r=.15 
\end{array} \right. 
\end{equation} 
for all values of $gV_{th}/I$ comprised between .9 and 1.25. Hence, the relative fluctuations cannot be neglected when $r=.15$. Figure \ref{fig-coupled}B shows the inference error  obtained from the Moving Threshold algorithm as a function of the membrane conductance for that value of the noise ratio. Not surprisingly, the Moving Threshold procedure is more accurate than the Fixed Threshold algorithm. 
 
In the Moving Threshold algorithm, the optimal potential is
constrained to remain below a certain threshold, $V^M_{th}$, which
depends on the time preceding the next spike and on the effective
current $I_i^e$.  Figure \ref{fig-mb}E shows the values of the moving
threshold $V^M_{th}$ and of the optimal potential $V_i^*$ for a few
spike intervals of the same neuron as in Fig.~\ref{fig-mb0}D. As
expected, the value of $V^*_i(t)$ lies substantially further away from
the threshold $V_{th}$ than in the Fixed Threshold procedure. In
addition, Fig. \ref{fig-mb}E shows a random realization of the
potential $V_i(t)$, obtained through numerical integration of the LIF
differential equation (\ref{ode}), for the same neuron $i$.  Although
$V_i$ is stochastic, the comparison of several inter-spike intervals indicates that $V^*_i(t)$ and $V_i(t)$ are in fair statistical agreement. 

To investigate in more details the origin of the inference error on
the couplings for large values of $r$ and $gV_{th}/I$, we plot in
Fig.~\ref{fig-negpos}C the inferred values of the interaction
$J_{ij}$ vs. the true value for every pairs $j\to i$ of a randomly
drawn network of $N=40$ neurons. The interactions inferred by the
Fixed Threshold algorithm are about ten times smaller than their true
values (Fig.~\ref{fig-negpos}C-3). The use of the Moving Threshold
procedure leads to a spectacular improvement for positive-valued
couplings (Fig.~\ref{fig-negpos}C-4).  While positive couplings are
accurately inferred, the magnitude of negative couplings is often
overestimated. These negative couplings are responsible for most of
the error $\epsilon _s$ in Fig.~\ref{fig-coupled}B.  From the Bayesian
point of view, when $\tau$ is smaller than the average ISI,
negative-valued couplings are indeed intrinsically harder to infer
than positive-valued ones.  A positive input drives the potential
closer to the threshold, which strongly reduces the ISI. Conversely, a
negative input drives the potential down, and a spike is unlikely to
occur before the potential first relaxes to its average level $I/g$
after a time of the order of $\tau$. Hence, the influence of a
negative input is hardly seen in the increase of the ISI when $\tau$
is smaller than the average ISI. We present an analytical calculation
supporting this argument in Section~\ref{sec-inf-neg}.

\subsection{Applications to multi-electrode recording data}
\label{sec-real}

We now apply our algorithm to multi-electrode recordings of the
ganglion cell activity of the salamander retina. Two data sets were
considered. The first one, hereafter referred to as Dark (data
courtesy of M. Meister), reports the spontaneous activity of 32
neurons for 2,000 seconds, and consists of $65,525$ spikes (Schnitzer
and Meister, 2003). In the second experiment, referred to as Natural
Movie (data courtesy of M. Berry), a retina was presented a 26.5
second-long movie, repeated 120 times, and the activity of 40 neurons
was registered for the whole duration of 3,180 seconds (Schneidman,
Berry, Segev and Bialek, 2006).  Natural Movie includes $172,521$
spikes. The firing rates, averaged over the population of recorded
neurons, have similar values in the two data sets: $f\simeq 1.02$
spikes/sec in Dark, $f\simeq 1.35$ spikes/sec in Natural Movie.

These two data sets were analyzed in a previous work (Cocco, Leibler and Monasson, 2009) with the perfect integrator model ($g=0$) and the Fixed Threshold algorithm. In this section we extend the analysis to the case of the LIF model and use both the Fixed and Moving Threshold approaches. In particular we show that the LIF model is  capable of inferring the asymmetry of the interactions, which is seen in the cross-correlograms but was not obtained with the perfect integrator model.
Moreover
 we discuss error bars on the inferred couplings and the  fact that strong
negative interactions are more difficult to infer than positive-valued
couplings.
We stress that the couplings we infer {\em a priori} depend on the stimulus.
Cocco, Leibler and Monasson (2009) have studied how the interactions 
inferred with the perfect integrator model depended on the stimulus based on the
analysis of two recordings on the same retina, namely the spontaneous
activity and random flickering squares. An alternative approach to disentangle stimulus-induced and structural contributions to the couplings would be to consider a time- and stimulus-dependent external current $I(t)$ (Section \ref{realistic}).

The value of the  membrane leaking time $\tau$ strongly affects the number of contacts and the running time of the algorithm. It takes about 40 seconds to infer
the currents and the interactions from either Dark or Natural Movie
when $\tau \simeq 1$ sec with one core of a 2.8 GHz Intel Core 2 Quad
desktop computer, and about 10 times longer when $\tau =100$ msec. The number of passive contacts of the optimal potential computed
by the Fixed Threshold procedure quickly decreases as $\tau$ increases. 
It is divided by $\simeq 20$ when the membrane leaking time increases
from $100$~msec to $10$~sec for both data sets. In comparison, the
number of active contacts is less sensitive to the value of $\tau$.
We find that the ratio of the number of contacts per neuron and per
second over the average firing rate takes similar values for both data
sets. For $\tau=1$ msec, this ratio is $\simeq 2.00$ for Dark, and
$\simeq 2.04$ for Natural Movie.
The number of passive contacts is smaller with the Moving Threshold
algorithm, while the number of active contacts remains rather
unchanged compared to its value with the Fixed Threshold procedure. 
On the overall, the running time of the Moving Threshold
procedure is higher due to the calculation of the time-dependent
threshold $V_{th}^M$.

Knowledge of the variance of the noise is required for the Moving
Threshold algorithm. The value of $\sigma$ could, in principle, be determined from experimental measures of the fluctuations of the synaptic current, but is unknown for the two recorded data sets available to us. We choose $\sigma$ so that the relative fluctuations of the
potential around the optimal path $V_i^*$ are less than $10\%$. We
compute these fluctuations by averaging (\ref{result-fluctu}) over all
ISI and all neurons $i$ in the population. The corresponding value of
the dimensionless standard deviation of the noise (\ref{defnounit})
are: for Dark, $\bar\sigma=.13, .12, .11$ for, respectively,
$\tau=200,100,20$ msec; for Natural Movie, $\bar\sigma=.15,.14,.12$
for, respectively, $\tau=200,100,20$ msec.

\subsubsection{Amplitudes  of the inferred interactions and currents}
\label{sec-amplitude}

\begin{figure}
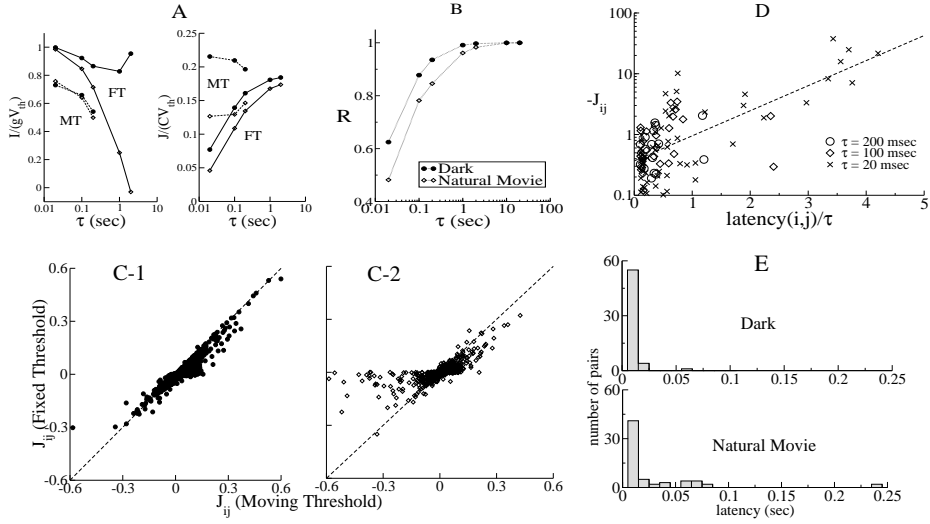

\begin{center}
\epsfig{file=fig5a.eps,width=4cm,height=3cm}
\hskip .2cm
\epsfig{file=fig5b.eps,width=3cm,height=3cm}
\hskip .2cm
\epsfig{file=fig5d.eps,width=4.5cm,height=3cm}
\vskip .3cm
\epsfig{file=fig5c.eps,width=7cm}
\hskip .2cm 
\epsfig{file=fig5e.eps,width=4.2cm,height=3.5cm}
\caption{Amplitudes of the interactions and currents in Dark (full circles) and
Natural Movie (empty diamonds). 
{\bf A.} Average value of the current (left) and root
mean square value of the coupling (right) as a function of the membrane
leaking time $\tau$. Points corresponding to the Fixed Threshold (FT)
procedure are joined by full lines, while dashed lines indicate the
results from the Moving Threshold (MT) algorithm.  Note that the
currents are larger in Dark than in Natural Movie.  {\bf B.}
Correlation coefficient $R$ (\ref{defcoeffrgeneral}) between the
couplings at leaking time $\tau$ and with no leakage.  {\bf C.}
Comparison between the interactions $J_{ij}$ found with the Moving
(x-axis) and the Fixed (y-axis) Threshold procedures, for Dark (C-1)
and Natural Movie (C-2). The dashed line is the $x=y$ line, and 
$\tau=20$ msec. 
{\bf D.} Strongly negative couplings $J_{ij}$ vs. latency
over the membrane leaking time $\tau$ for three values of
$\tau$. Couplings were obtained using the Moving Threshold procedure,
and correspond to the Natural Movie data set. Only interactions
$J_{ij}<-.1$ are considered; there are, respectively, 16, 28, and 60
such couplings for $\tau=200$, 100, and 20 msec. The value of the
slope of the best linear fit $\log (-J_{ij}) = \alpha$
latency$(i,j)/\tau+\beta$, shown by the dashed line, is $\alpha=0.95$.
{\bf E.} Distributions of the latencies (\ref{deflat}) between neurons
in Dark (top) and Natural Movie (bottom). Only latencies larger than 5
msec are taken into account in the histograms.}
\label{fig-amplitude}
\label{fig-compar-J}
\label{fig-fm}
\label{fig-scaling-tau}
\label{fig-latency}
\end{center}
\end{figure}

Figure \ref{fig-amplitude}A shows the average value of the current and
of the interaction strength as a function of the membrane leaking
time. As expected with the Fixed Threshold inference procedure, we
find that the average value of the couplings decreases as $\tau$ gets
small. This effect varies from neuron to neuron: the closer $I_i$ is
to $gV_{th}$, the smaller are the couplings $J_{ij}$.  To compare the
matrices of couplings $J,J'$ inferred with the Fixed Threshold
algorithm for different values of $\tau$, we use the correlation
coefficient (Hubert and Baker, 1979)
\begin{equation}\label{defcoeffrgeneral}
R(J,J') = \frac{\hbox{\rm cov}(J,J')}
{\sqrt{ \hbox{\rm cov}(J,J)\; \hbox{\rm cov}(J',J')}} \ ,
\end{equation}
where
\begin{equation}
\hbox{\rm cov} (J,J') = N(N-1) \sum_{i\ne j} J_{ij}\, J'_{ij} - 
\big( \sum _{i\ne j} J_{ij}\big)\big( \sum _{i\ne j} J'_{ij}\big) \ .
\end{equation}
Identical matrices correspond to $R=1$, and uncorrelated matrices give $R=0$. 
$R$ is independent of the scale of the coupling matrices $J$ and $J'$,
{\em i.e.}  $R(a J,a'J')=R(J,J')$ for any $a,a'> 0$; therefore, $R$ is sensitive to the relative amplitudes of the couplings $J'$ and $J$ and not to their absolute differences. We
choose $J$ to be the coupling matrix in the absence of leakage and
$J'$ to be the coupling matrix for a given $\tau$. The value of $R$ as
a function of $\tau$ is shown in Fig.~\ref{fig-compar-J}B.  Even for
$\tau=20$ msec, the coupling matrix is substantially similar to the
one obtained with the perfect integrator model ($R=.6$ for Dark,
$R=.5$ for Natural Movie). Despite the overall change in the amplitude
of the inferred couplings, the relative ordering of the couplings with
the pair indices $(i,j)$ is largely independent of $\tau$, especially
so for Dark.  However, for specific pairs of neurons, the interactions
may strongly depend on $\tau$. Such a dependence effect will be
illustrated in Section \ref{sec-symmetry}, and can be related to the
temporal structure of the corresponding cross-correlograms.

The average value of the interactions calculated by the Moving
Threshold algorithm does not decrease when $\tau$ gets smaller, and is
larger than the one obtained from the Fixed Threshold procedure
(Fig.~\ref{fig-amplitude}A).  To better understand this discrepancy,
we compare in Fig.~\ref{fig-fm}C the interactions inferred with both
algorithms for every pairs of neurons in the Dark and Natural Movie
data sets when $\tau = 20$ msec.  The agreement between both
procedures is very good for positive and strong couplings. Couplings
which are slightly positive with the Fixed Threshold procedure generally
have a larger value with the Moving Threshold procedure. This offset
is responsible for the differences in the average values of the
interactions found in Fig.~\ref{fig-amplitude}A.  In addition, in
Natural Movie, negative-valued couplings often have a stronger
amplitude with the Moving Threshold procedure. We find, in both
approaches, a few negative and very strong couplings. The amplitude of
those extreme couplings increases very quickly as the membrane leaking
time decreases.

The emergence of strong negative interactions with the lowering of
$\tau$ can be related to the presence of long latencies between the
emission of spikes. We define the latency of neuron $i$ with respect
to neuron $j$ as the smallest delay between a spike emitted by $j$ and
a later spike fired by $i$,
\begin{equation}\label{deflat}
\hbox{\rm latency}(i,j) = \min _{k,\ell: t_{i,k} < t_{j,\ell} < t_{i,k+1}} (t_{i,k+1}-t_{j,\ell}) 
\end{equation}
A large value of the latency of neuron $i$ with respect to $j$ is interpreted by the 
inference procedure as the consequence of a strongly inhibiting coupling from
$j$ to $i$. However, the effect of a synaptic input of amplitude $J_{ij}$ on the 
potential $V_i$ of the neuron $i$ decays exponentially with the ratio of the time 
elapsed from the input over the membrane leaking time. Hence, to keep the latency fixed
while $\tau$ is changed, the strong and negative interaction must change accordingly,
\begin{equation} \label{jtau}
J_{ij} \sim  \hbox{
\rm constant} \times \exp\left( \frac{\hbox{\rm latency}(i,j)}{\tau}\right) \ ,
\end{equation}
where the constant has a negative value. Figure~\ref{fig-scaling-tau}D 
shows the negative couplings $J_{ij}$ vs. the
latencies of the corresponding pairs $(i,j)$ divided by 
$\tau$, for three values of $\tau$. The outcome suggests 
that relation (\ref{jtau}) is indeed correct, see Fig.~\ref{fig-scaling-tau}D and its
caption. 

The above mechanism explains why strongly negative
couplings are less frequent in Dark than in Natural Movie.  For $\tau=100$ 
msec, there are 10 interactions (out of 1560) smaller than $-1$ in Natural Movie,
and none (out of 992) in Dark.  For $\tau=20$ msec, these two numbers
are equal to, respectively, 23 and 1. Figure \ref{fig-latency}E shows the 
histograms of latencies for both data sets. In Natural Movie,
we find 17 pairs with latencies larger than 25 msec.
In Dark, only one pair $(i,j)$ has a latency larger than 25 msec. The corresponding
interaction, $J_{ij}$, is the only one smaller than $-1$ for $\tau=20$ msec. 

\subsubsection{Accuracy on the inferred interactions and currents}
\label{sec-inf-neg}

As discussed in the Methods section, the uncertainty on the inferred
parameters can be obtained from the Hessian matrix of $L^*$, that is, from the
curvature of the log-likelihood around its maximum. To quantify those uncertainties,
we use the following procedure. Assume for instance we want to know how reliable is the
inferred value, $\hat J_{i,j_0}$, of the interaction $J_{i,j_0}$ from neuron
$j_0$ to neuron $i$. We fix $J_{i,j_0}$ to
an arbitrary value, and maximize $L^*({\cal T}|{J_{ij}}, I_i)$ (\ref{deflpcc}) 
over all the couplings $J_{ij}$ with $j\ne j_0$ and over the current $I_i$ . 
The outcome is a function of $J_{i,j_0}$, which we denote by $L_c$ and 
call marginal log-likelihood. 
$L_c(J_{i,j_0})$ has, by definition, a maximum in $J_{i,j_0}=\hat J_{i,j_0}$. 
Its second derivative in the maximum, $L_c''(\hat J_{i,j_0})$, is related to the error bar
$\Delta J_{i,j_0}$ on the interaction through, see (\ref{fluch}) and (\ref{flucov}),
\begin{equation}\label{eb09}
\Delta J_{i,j_0} = \sqrt{\langle (J_{i,j_0} -\hat J_{i,j_0})^2\rangle} =  
\frac{\sigma} {\sqrt{-L_c''(\hat J_{i,j_0})}} \ .
\end{equation}
The same procedure can obviously be used to obtain the error bar on the current $I_i$.

\begin{figure}
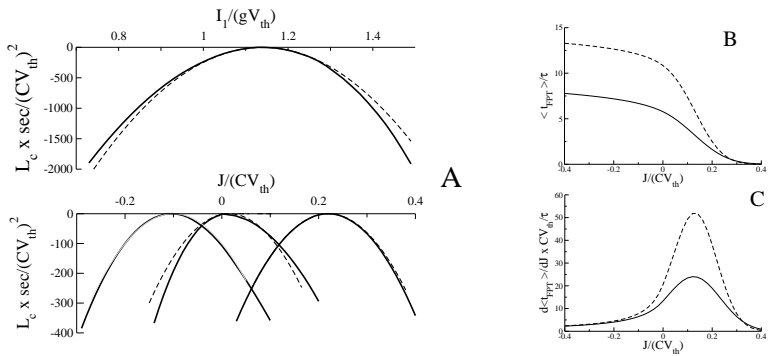

\begin{center}
\hskip -2cm
\epsfig{file=fig6a.eps,width=6cm}
\vskip -4.2cm
\hskip 9.cm
\epsfig{file=fig6b.eps,width=3.cm}\\
\hskip 9.cm
\epsfig{file=fig6c.eps,width=3.cm}
\caption{{\bf A.} Marginal log-likelihoods $L_c(I_1)$ (top panel), and $L_c(J_{1,27})$, 
$L_c (J_{1,20})$,  
$L_c(J_{1,4})$ (from left to right in the bottom panel) for Natural Movie,
and $\tau=200$~msec. 
Dashed lines correspond to the best fits with a single quadratic function. 
The most likely value for the current is $\hat I_1=1.14 gV_{th}$. 
The most likely values for the interactions are: 
$\hat J_{1,27}=-.11$, $\hat J_{1,20}=.01$, and $\hat J_{1,4}=.22$, in units of $CV_{th}$.
The offset on the vertical axis has been
chosen so that all maxima are at height $L_c=0$. 
{\bf B.} Average value of the first-passage time $t_{FPT}$ 
after a synaptic entrance of amplitude $J$. 
{\bf C.} Derivative of $t_{FPT}$ with respect to
$J$. The parameters of the neuron are: $gV_{th}/I=1.2$, $r=.15$, 
$\tau=85$ msec (full line) 
and 20 msec (dashed line). The derivative is maximal 
around $J_{opt}/(CV_{th})=1-I/(gV_{th})\simeq .167$.}
\label{fig-curvature}
\label{fig-tfpt}
\end{center}
\end{figure}

We now illustrate this approach on the Natural Movie data set, and one 
arbitrarily chosen neuron, $i=1$. Three interactions,
representative of, respectively, positive, weak, and negative couplings, 
were singled out among the 39 couplings incoming onto neuron~1.
Figure \ref{fig-curvature}A shows the marginal log-likelihoods 
$L_c(J_{1,4})$, $L_c (J_{1,20})$, and $L_c(J_{1,27})$, in addition to $L_c(I_1)$.
For all four parameters, the marginal likelihoods can be approximated with parabolas 
in the vicinity of their maxima. Estimating the second derivatives from those
best quadratic fits and using (\ref{eb09}), we obtain
\begin{equation}
\frac{\Delta I_1}{gV_{th}} \simeq .020\ \bar \sigma\ ,\quad 
\frac{\Delta J_{1,27}}{CV_{th}} \simeq .023\ \bar \sigma\ ,\quad 
\frac{\Delta J_{1,20}}{CV_{th}} \simeq .021 \ \bar \sigma\ ,\quad 
\frac{\Delta J_{1,4}}{CV_{th}} \simeq .022 \ \bar \sigma\ .
\end{equation}
where $\bar\sigma$ is the dimensionless noise level defined in (\ref{defnounit}).
Hence, the error bars on the couplings and currents have very similar values.
This common value depends on the noise level, $\bar \sigma$. As discussed in the
next Section~\ref{sec-amplitude}, $\bar \sigma$ is expected to be
close to, or smaller than unity when $\tau=200$ msec. Consequently, the value for
$J_{1,20}$ is compatible with zero, while the interactions
$J_{1,27}$ and $J_{1,4}$ are non zero, with 99.9999\% confidence.

A closer inspection of Fig.~\ref{fig-curvature}A shows that the
quality of the quadratic fit of $L_c$ is excellent for $J_{1,27}$ and
$J_{1,4}$, but less so for $I_1$ and $J_{1,20}$.  For the latter
parameters, it seems that the curvature of $L_c$ takes two different
values, depending on whether the maximum is approached from the left
of from the right. This phenomenon results from the piece-wise
structure of the $L^*$ function, see Methods section. A practical
consequence is that the errors $I_1-I_1^*$ and $J_{1,20}-\hat
J_{1,20}$ are not evenly distributed around zero; for instance
$J_{1,20}$ is more likely to be larger than $\hat J_{1,20}$ than it is
to be smaller.

Note that strong, negative interactions may be harder to infer than
positive-valued couplings, a phenomenon already underlined by Aersten
and Gerstein (1985). The underlying intuition is that the duration of
the ISI is less affected by an inhibitory input than by an excitatory
input when the membrane leaking time, $\tau$, is small compared to the
average value of the ISI.  We now present an analytical argument
supporting this intuition.  Consider a neuron, fed with an external
current $I$ and with noise variance equal to $\sigma^2$.  Assume a
synaptic input of amplitude $J$ is received at time $t=0$. We call
$t_{FPT}$ the average value of the time at which the neuron will emit
a spike; the calculation of $t_{FPT}$ can be done using a series of parabolic cylinder functions (Alili, Patie and Pedersen 2005). Figures \ref{fig-tfpt}B\&C shows that the dependence of $t_{FPT}$ on $J$ is much weaker for negative-valued $J$ than for
positive couplings.  As the set of spiking times is the only
information we have at our disposal, the difficulty in inferring
negative couplings is intrinsic to the Bayesian approach, and cannot
be circumvented by any particular algorithm.

\subsubsection{Symmetry of the interactions and cross-correlograms}\label{sec-symmetry}

\begin{figure}
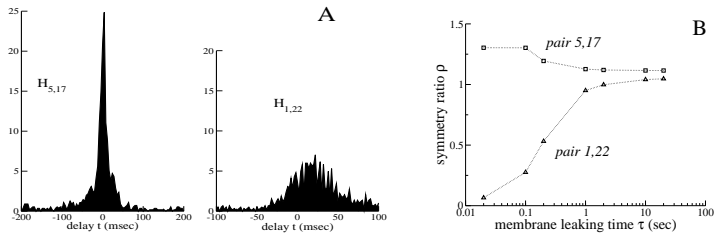

\begin{center}
\epsfig{file=fig7a.eps,width=5cm,height=3.cm}
\hskip .5cm
\epsfig{file=fig7b.eps,width=3.7cm}
\caption{{\bf A.} Cross-correlograms $H(t)$ for pairs $(5,17)$ and $(1,22)$ in Dark.
The cross-correlograms are normalized such that $H(t)\to 1$ for large delays $|t|$. 
{\bf B.} Ratios $J_{ij}/J_{ji}$ of the interactions between the neurons 5,17
(top) and $1,22$ (bottom) as a function of $\tau$. }
\label{fig-rapportj}
\label{fig-coefficient_r}
\label{fig-symmetry}
\end{center}
\end{figure}

The dependence of the symmetry of couplings upon the membrane leaking time $\tau$ can be understood, to some extent, from the structure of the cross-correlograms, that is, the histograms $H_{ij}(t)$ of the delays $t=t_{i,k}-t_{j,\ell}$ between the times of the spikes fired by the two neurons $i,j$ in each pair.  To do so, we consider two pairs of neurons in Dark, called pairs $(5,17)$ and $(1,22)$. Figure \ref{fig-rapportj}A shows the cross-correlograms $H_{5,17}$ and $H_{1,22}$. Pair $(5,17)$ is characterized by a positive peak in $H$, centered in $t=0$, and of width $\simeq 20$ msec. Pair $(1,22)$
exhibits a positive peak of correlations, of the same width, but centered around $t\simeq 20$~msec.

We plot in Fig.~\ref{fig-rapportj}B the symmetry ratios of the interactions in the pairs, $\rho_{5,17}=J_{5,17}/J_{17,5}$ and $\rho_{1,22}=J_{1,22}/J_{22,1}$. We find that $\rho_{5,17}$ is, to a large extent, independent of $\tau$. Conversely, $\rho_{1,22}$ sharply
decreases with decreasing $\tau$ and is close to zero when $\tau=20$
msec, which coincides with the typical delay in the cross-correlogram
$H_{1,22}$ shown in Fig.~\ref{fig-rapportj}A. We conclude that the
inference procedure is capable of capturing the directionality of the
interaction between the neurons 1 and 22, if $\tau$ is small
enough. This results shed some light on the correspondence between the
interactions inferred within the LIF model and within the Ising model
(Schneidman, Berry, Segev and Bialek, 2006; Shlens et al, 2006).
Couplings inferred with the perfect integrator model for Dark are in
good agreement with the Ising interactions, when the time is binned
into windows of width $\Delta t=20$ msec (Cocco, Leibler and Monasson,
2009).  By construction, the Ising model produces symmetric
interactions from the pair-wise correlations of the activities,
averaged of the binning window. In the absence of leakage, the
Integrate-and-Fire inference algorithm hardly distinguishes between a
post-synaptic and pre-synaptic firing pattern, and produces rather
symmetric couplings. But as $\tau$ decreases, the LIF couplings may
become strongly asymmetric (Fig.~\ref{fig-rapportj}B). In this case,
the correspondence between the Ising and LIF couplings breaks
down. The same phenomenon was observed in Natural Movie, where delays
in the cross-correlograms are even stronger.

\section{Discussion}\label{conclusions}

In this article, we have presented a procedure to infer the
interactions and currents in a network of Leaky Integrate-and-Fire
neurons from their spiking activity. The validity of the procedure was established
through numerical tests on synthetic data generated from networks with
known couplings. We have also applied our algorithm to real
recordings of the activity of tens of ganglion neurons in the salamander 
retina. Though our algorithm is limited to moderate noise levels and instantaneous 
synaptic integration, it is 
fast and can, to our knowledge, handle much bigger data sets than the existing
inference methods for the stochastic IF model. It is our intention to
make this algorithm available to the neurobiology community in a near
future.

\subsection{Comparison with previous studies}

Cross-correlation analysis (Perkel, Gerstein and Moore, 1967; Aersten
and Gerstein, 1985) consists in studying the distribution of delays
between the spikes of neurons in a pair. This approach has been used
to characterize the connections between neurons (amplitude,
time-scale, dependence on distance), or their dynamical evolution
(Fujisawa, Amarasingham, Harrison and Buzsaki, 2008).  The analysis do
not require any combinatorial processing of the activity of a large
part of the neural assembly. As a result, the approach is not limited
to small networks. However, cross-correlation analysis may find
difficult to separate direct correlations from indirect correlations
modulated through interactions with neurons in the surrounding network
(Ostojic, Brunel and Hakim, 2009; Cocco, Leibler and Monasson, 2009), or due to common inputs
(Constantidinidis, Franowicz and Goldman-Rakic, 2001; Trong and Rieke,
2008).

In statistical approaches a widely-used concept is the one of
causality (Seth and Edelman, 2007). A causal interaction exists from
neuron $i$ to neuron $j$ if the knowledge of the activity of $i$ helps
predict the firing of $j$ beyond what can be achieved from the
activity of $j$ alone.  In practice, causal relationships are detected
through linear multivariate statistical regressions (Sameshima and
Baccal\'a, 1999), and may overlook non-linear dependencies. Causal
analysis have also difficulties in evaluating the strength of the
interactions.

Maximum entropy models, which deduce interactions from pairwise
correlations only, have been shown to accurately reproduce
higher-order correlations between neurons in the vertebrate retina
(Schneidman, Berry, Segev and Bialek, 2006; Shlens et al, 2006; Cocco,
Leibler and Monasson, 2009). These models, however, suffer from some
limitations. Interactions are constrained to be symmetric, and
temporal correlations are partially discarded (Marre, El Boustani,
Fr\'egnac and Destexhe, 2009).  In addition obtaining the interactions
from the correlations may be computationally very hard for large
networks, though efficient approximate
algorithms have recently been developed (Cocco and Monasson, 2010).

Generalized linear models (GLM), which represents the generation of spikes
as a Poisson process with a time-dependent rate, have been applied to
various neural systems (Brown, Nguyen, Frank, Wilson and Solo, 2001; 
Truccolo et al, 2005; Pillow et al, 2008). The inference of parameters
in the GLM framework is apparently easier to solve than for IF models, which
has made the GLM framework very attractive. Whether GLM are better than IF 
models to account for real neural activity, regardless of the computational 
complexity of both inference framework, is an important issue
(Gertsner and Naud, 2009). We hope that our work, which
makes possible to apply the IF model to large data sets, will 
help to answer this question.

Approaches to infer model parameters in the IF framework have been so far capable of processing a very limited number of neurons or of spikes. Pillow et al. (2005) inferred the model parameters of one stochastic IF neuron based on a 50 second-long recording with a procedure tolerating any level of noise; Makarov, Panetsos and de Feo (2005)
inferred the connections between 5 deterministic IF neurons from a 60 second-long synthetic spike train. In comparison we have analyzed a 3180-second long recording of the activity of 40 neurons.

The running time of our procedure increases as $N^2\; S \sim N^3\; T\,f$, where $T$ is the duration of the recording and $f$ is the average firing rate. Recently, Koyama and Paninski (2009) have proposed a numerical procedure for calculating the optimal potential and inferring the interactions. In their approach, the time is discretized into many time-bins of small duration $\Delta$, and the values of the optimal potentials at those discrete times can be found by means of the interior-point method for discrete constrained optimization problems.  The running time of the procedure, $O(N^3 \;T/\Delta)$, is approximately $1/(f\Delta)$ times larger than ours. In practice, $f$ is of the order of 1 to 10 Hz, while the discretization time, $\Delta$, is of the order of 1 msec; hence, $1/(f\Delta)$ ranges from 100 to 1000. However, this order of magnitude does not take into account the existence of multiplicative constants; a comparative test of the two approaches on the same synthetic or real data would be useful to accurately estimate their running times. Furthermore, the algorithm introduced by Koyama and Paninski can easily incorporate the presence of temporal filtering in the interactions. Our procedure is, in its present form, valid when the integration kernel is instantaneous only; considering other synaptic kernels would require {\em ad hoc} modifications to the expressions of the optimal noise and potential and to the search procedure for contacts.

\subsection{How to include a finite integration time}

One of the major assumptions in our approach is that the synaptic
integration time, $\tau _s$, is vanishingly small. In practice, $\tau
_s$ does not vanish, but might often be smaller than the membrane leaking 
time, $\tau$, and the average ISI.  Assume that neuron $i$, whose potential
$V_i$ is close to the threshold $V_{th}$,
receives a spike at time $t$ from another neuron, $j$, through a strongly excitatory
connection $J_{ij}> V_{th}-V_i$. Then, neuron
$i$ will reach the threshold level after having received a charge
$\Delta q = C(V_{th}-V_i)$, smaller than $J_{ij}$. As a
consequence, large positive interactions can be underestimated  when 
the latency of neuron $i$ from neuron $j$ (\ref{deflat}) is smaller than
$\tau_s$. 

To compensate for this effect we could introduce a time-dependent value for the 
interaction,
\begin{equation}
J_{ij} (t,t_{i,k+1}) = J_{ij}  \; \min \bigg( \frac {t_{i,k+1}-t}{\tau_s} ,1
\bigg) \ ,
\end{equation}
where $t_{i,k+1}(>t)$ is the closest firing time of neuron $i$. 
Hence the effective interaction $J_{ij} (t,t_{i,k+1})$ is equal to its nominal value 
$J_{ij}$ only if the synaptic current has enough time to enter the neuron $j$, and
is a fraction of $J_{ij}$ otherwise.
The modified procedure will be correct as long as $\tau_s<\tau$. 
If the synaptic and membrane time-scales are 
comparable, one needs to take into account the complete shape of the
synaptic integration kernel, $K(t)$.
Choosing simple enough integration synaptic kernel, such as
the piece-wise linear function $K(t)=0$ if $t<0$ or $t>\tau_s$,
$K(t)=2 \min (t,\tau_s-t)/\tau_s^2$ if $0\le t\le \tau_s$, could
lead to tractable dynamical equations for the optimal potential and noise.
The resolution of those equations is left for future work.

\subsection{Towards a more realistic inference model}

The inference procedure that we have introduced here can be extended
to include realistic features such as a refractory period, $\tau_R$. To do so,  
we restrict the sum in (\ref{defv}) to the spikes $m$ entering the neuron $i$ at times
larger than $t_0+\tau_R$.  We have run the modified inference
procedure on the recordings of the retinal activity, for values of
$\tau_R$ ranging from 2 to 5 milliseconds.  The couplings did not
change much with respect to the values found with $\tau_R=0$. Note that the introduction of a propagation delay $\tau_{D}$ in the synaptic interaction is straightforward, as long as the integration kernel remains a Dirac distribution (centered in $\tau_{D}$). 

Bounds on the values of the couplings and currents {\em e.g.} to
prevent the exponential growth of negative interactions with the
leaking conductance can naturally be introduced through a prior
distribution. As an example, assume that the interactions $J_{ij}$
take values in $[J_-,J_+]$.  Then, one could maximize
$L^*-\displaystyle{\sum _{i,j}} W(J_{ij})$ instead of the
log-likelihood $L^*$ alone, where $W(J)=\frac w2 (J-J_-)^2$ if
$J<J_-$, 0 if $J_-<J<J_+$, $\frac w2 (J-J_+)^2$ if $J>J_+$ and $w$ is
a large positive coefficient.

We have assumed, throughout this work, that the values of $g$ and $V_{th}$ were known. In practical situations, while the orders of magnitudes are known, the precise values of these parameters should be inferred, and could depend on the neuron $i$. The inference procedure could be modified to update the values of $g_i$ and $(V_{th})_i$ at the same time as the synaptic couplings $J_{ij}$ and the current $I_i$. The number of parameters to infer (per neuron) would simply increase from $N$ to $N+2$, and the running time should not increase too much.

\subsection{Inference from a limited neural population and in the presence of a stimulus} 
\label{realistic}

Nowadays, multi-electrode experiments can record a few tens,
or hundreds of neurons. To which extent do the interactions inferred 
from this sub-population coincide with the interactions one would
find from the knowledge of the whole population activity? 
The question does not arise in cross-correlation analysis: the correlation
between the firing activities of two neurons is obviously independent 
of whether a third neuron is recorded or not. However the issue 
must be addressed as soon as a collective model for generating 
the activity is assumed, such as the coupled LIF models studied here.  

A detailed analysis suggests that the interaction between a pair of neurons is not affected by the activity of other neurons distant by more than $\ell=300\ \mu$m in the case of spontaneous activity (Cocco, Leibler and Monasson, 2009). The electrode array should be at least twice longer and wider than $\ell$, and should be dense enough to capture all the neurons on the recorded area. It is estimated that about 10\% of the ganglion cells are registered in the Dark experiment, compared to more than 80\% with the 
denser but smaller electrode array used in the Natural Movie experiment  (Segev, Puchalla and Berry, 2005). It would thus be very interesting to repeat our study on other multi-electrode recordings, with sufficiently large and dense arrays. 

Taking into account the stimulus $S$ in the inference process would also be interesting. To do so, we could add a stimulus-induced current, $I^{s}(t|S)$, to (\ref{ode}). A simple expression for this current would be $I^{s}(t|S)=\int_0^t {dt'} K_i^{s} (t-t') \; S _i (t')$, where $K_i^s$ is a kernel similar to the one used in generalized linear models (Pillow et al, 2008). The expression of the current-dependent term in the potential $V(\eta,t)$ (\ref{defv}) should be modified accordingly, while the noise-dependent term would remain unchanged. It is important to note that the search procedure for contacts presented in Section \ref{sec-algo} would remain valid. However, the expressions of the noise coefficient, the contact time and the duration of a passive contact given in Appendix \ref{secgnon} for the case of a constant current $I$ should be rederived and would depend on the precise temporal structure of the stimulus-induced current $I^s(t|S)$.  

\vskip 1cm
\noindent{\bf Acknowledgment:} 
This work originates from a collaboration with S. Leibler, whom we
thank for numerous and fruitful discussions. We thank C. Barbieri for
a critical reading of the manuscript. We acknowledge the
hospitality of The Rockefeller University, where this work was
initiated. Partial funding was provided by the Agence Nationale
de la Recherche under contract 06-JCJC-051.

\appendix

\section{Active contacts}\label{defsec2}\label{seccp}

In this Appendix, we justify the prescriptions in the search for active contacts
presented in Section \ref{sec-algo}. For the sake of simplicity we 
restrict to the $g=0$ case (no membrane leakage); the extension to non-zero
$g$ is briefly discussed in Appendix \ref{secgnon}. 
We consider a neuron $i$, and call $M$ the number of spikes received 
by this neuron during its $k^{th}$ inter-spike interval 
$[t_0\equiv t_{i,k};t_{M+1}\equiv t_{i,k+1}]$. The arrival times
are $t_1<t_2< \ldots < t_M$, and the corresponding synaptic strengths
are $J_1,J_2,\ldots , J_M$. To lighten notations we hereafter omit the index $i$ of
the neuron. 

\subsection{Case of $M=0$ or 1 input spike}
\label{illustration}

\begin{figure}
\begin{center}
\epsfig{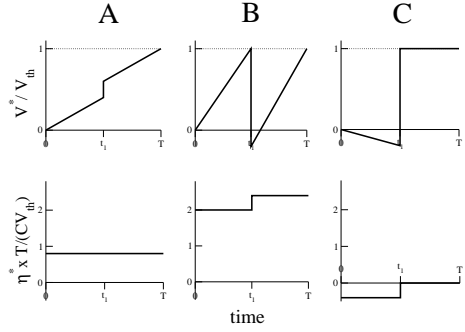}
\caption{Sketches of the optimal potentials $V^*$ (top) and noises $\eta^*$ 
(bottom) for one neuron receiving one weak ({\bf A}), one strong negative 
({\bf B}), and one strong positive ({\bf C}) input. The jump in the optimal noise consecutive to an active contact is always positive. Values of the parameters used for the figure are: $I=0$, $t_1=T/2$, $J_1/(CV_{th})=.2$ ({\bf A}), -1.2 ({\bf B}), 1.2 ({\bf C}).}
\label{fig-sol1}
\end{center}
\end{figure}

To understand the key notion of contact, we first consider the simple case of a neuron receiving no spike during the inter-spike interval $[t_{i,k}=0;t_{i,k+1}=T]$. The optimal noise is constant according to (\ref{2.8}). Equation (\ref{optimb}) then shows that  
the optimal potential is a linear function of the time, which is fully determined from the boundary conditions $V^*(0)=0,V^*(T)=V_{th}$. We obtain
\begin{equation} \label{illm0}
V^*(t) =V_{th}\; \frac{t}{T}  \qquad \mbox{and} 
\qquad  \eta ^* (t) = \frac {CV_{th}}{T} -I \ .
\end{equation}
This solution is correct since the potential remains below the threshold at all times $0<t<T$. 

Let us now assume now that the neuron receives one input from another neuron,
of strength $J_1$ at time $t_1 \in ]0;T[$. 
The effect of the input is a discontinuous jump of the potential
at time $t_1$ and of size $\frac{J_1}C$, shown in Fig.~\ref{fig-sol1}. 
Repeating the calculation above, we obtain 
the following expressions for the optimal potential and noise
\begin{equation}\label{illma}
 V_A^*(t) = \left(V_{th}-\frac{J_1}C\right)\frac{t}{T} + \frac{J_1}C\; \theta (t-t_1) \quad
\mbox{and} \quad 
\eta _A ^*  = \frac {CV_{th}-J_1}{T} -I   \qquad
(\mbox{case A})\  ,
\end{equation}
where $\theta$ is the Heaviside function: $\theta(x)=1$ if $x>0$, 0 otherwise.
This solution is sketched in Fig.~\ref{fig-sol1}A. It is valid when
the potential $V_A^*$ remains below the threshold at all times. We call 
this situation 
case A. As $V_A^*$ is a piece-wise linear function we only need to check that 
$V^*_A(t_1^-)$ and $V_A(t_1^+)$ are both smaller than $V_{th}$. The two conditions are
fulfilled provided that
\begin{equation}
J_- \equiv -CV_{th}\frac{T-t_1}{t_1} < J_1 < J_+ \equiv CV_{th} \ .
\end{equation}
What happens when the above condition is violated? Let us
consider first $J_1< J_-$ (referred to as case B hereafter). Then
$V_A^*$ exceeds the threshold $V_{th}$ before the input enters the
neuron. To prevent the potential from crossing the threshold at time $t_1$, 
the true optimal noise, $\eta _B^*$, should be 
smaller than $\eta^*_A$. But, if $\eta_B^*<\eta_A^*$, the
potential could not reach $V_{th}$ when the
neuron emits its spike at time $T$ according to the very definition of $\eta_A^*$! 
The only way out is that $\eta_B^*$
takes two different values corresponding to the two sub-intervals
$[0;t_1[$ and $]t_1;T]$, which we call, respectively,
$\eta_{B,-}^*$ and $\eta_{B,+}^*$. We expect $\eta_{B,-}^* < \eta_A^* <\eta_{B,+}^*$. 
The noise can change value in $t=t_1$  through 
(\ref{optimc}) only if the potential reaches the threshold
in $t_1$. We find that
\begin{equation}
V_B^*(t) = V_{th} \;\frac{t}{t_1}\qquad \mbox{and} 
\qquad  \eta_{B,-}^* =  \frac {CV_{th}}{t_1} -I   \qquad
(\mbox{case B}, 0<t< t_1)\  ,
\end{equation}
from the boundary conditions $V^*(0)=0,V^*(t_1^-)=V_{th}$, and
\begin{equation}
V_B^*(t) = V_{th} + \frac{J_1}C\;\frac{T-t}{T-t_1}  \qquad \mbox{and} 
\qquad  \eta_{B,+}^* = -\frac {J_1}{T-t_1} -I \qquad
(\mbox{case B}, t_1<t< T)\  ,
\end{equation}
from the boundary conditions $V^*(t_1^+)=V_{th}+\frac{J_1}C,V^*(T)=V_{th}$.
This solution is drawn in Fig.~\ref{fig-sol1}B. It is important to stress that 
the above solution is based on the capability of the noise to
abruptly change its value when the potential touches the threshold
in $t=t_1$. A detailed study of the behavior of the noise close to 
such `contact points' proving that this is indeed the case 
is postponed to Appendix \ref{seccp-sec}.

Finally, we turn to case C corresponding to $J_1>J_+$. In this case the 
input is so excitatory that the noise has to be negative to prevent the
neuron from emitting a spike at a time $t<t_1$. As in case B,  the
potential reaches the threshold in $t=t_1$ to allow the noise to
change its value after the input has entered the neuron. We find
\begin{equation}
V_C^*(t) = \left(V_{th}-\frac{J_1}C\right)\frac{t}{t_1} \qquad \mbox{and} 
\qquad  \eta_{C,-}^* =  \frac {CV_{th}-J_1}{t_1} -I  \qquad
(\mbox{case C}, 0<t< t_1)\  ,
\end{equation}
according to the boundary conditions
$V_C^*(0)=0,V_C^*(t_1^-)=V_{th}-\frac{J_1}C$. Right after the spike has been
received, the potential has reached its threshold value, and will
keep to this value until a spike is emitted at time $T$, hence
\begin{equation}
V_C^*(t) = V_{th}   \qquad \mbox{and} 
\qquad   \eta_{C,+}^* = -I\qquad
(\mbox{case C}, t_1<t< T)\  .
\end{equation}
This solution is drawn in  Fig.~\ref{fig-sol1}C. 

We now give the values of log-likelihoods $L^*$ corresponding to the cases listed above. The value of $L^*$ can be calculated from the knowledge of the optimal noise
$\eta^*$ through (\ref{deflpcc}). In the case of $M=0$ spike, we find, using
(\ref{illm0}) with $T=t_1-t_0$,
\begin{equation}
L^*(t_0,t_1|I)  = - \frac{(CV_{th}-I\, ({t_1-t_0}))^2}{2 ({t_1-t_0})} \ .
\end{equation}
The optimal current is then inferred by maximizing $L^*(I)$ with the
result $\hat I=\frac 1{t_1-t_0}$, which corresponds to a vanishing value for the
optimal noise, as expected.

When $M=1$ spike is received by the neuron, the log-likelihood $L^*$ has three
distinct expressions corresponding to the case A, B, C discussed in
Section \ref{illustration}. The resulting expression is (with $t_2=T$):
\begin{equation}
L^*(t_0,t_1,t_2|J_1,I) = \left\{ 
\begin{array} {c c c}
- \frac{(CV_{th}-J_1- I\, (t_2-t_0))^2}{2 (t_2-t_0)} & \mbox{if} & 
J_-< J_1 < J_+ \qquad \mbox{(case A)} \\
- \frac {(CV_{th}- I\, (t_1-t_0))^2}{2 (t_1-t_0)} - \frac{(J_1+ I\, (t_2-t_1))^2}{2
  (t_2-t_1)}  &  \mbox{if} & 
J_1 < J_- \qquad \qquad \mbox{(case B)} \\
- \frac {(CV_{th}-J_1- I\, (t_1-t_0))^2}{2 (t_1-t_0)}  -\frac {I^2}
2(t_2-t_1) &  \mbox{if} & 
J_1 > J_+ \qquad \qquad \mbox{(case C)} 
\end{array} \right. \ .
\end{equation}
The log-likelihood $L^*$ is a continuous and convex function of its
argument. The first derivatives of $L^*$ are continuous in $J_-,J_+$,
but the second derivatives are not.

\subsection{Study of the optimal noise close to an active contact point}
\label{seccp-sec}

The noise coefficient $\eta$ in (\ref{eq1}) are constant over the time
interval separating two active contacts. The value of $\eta$ may
however change upon the crossing of an active contact.  The scope of
this section is to show that the noise right after the contact can take {\em any value} larger than the noise immediately before the contact. This monotonicity property justifies the search for the minimal noise coefficient done in (\ref{defnoise}), see Appendix \ref{seccp2}.

\begin{figure}
\begin{center}
\epsfig{file=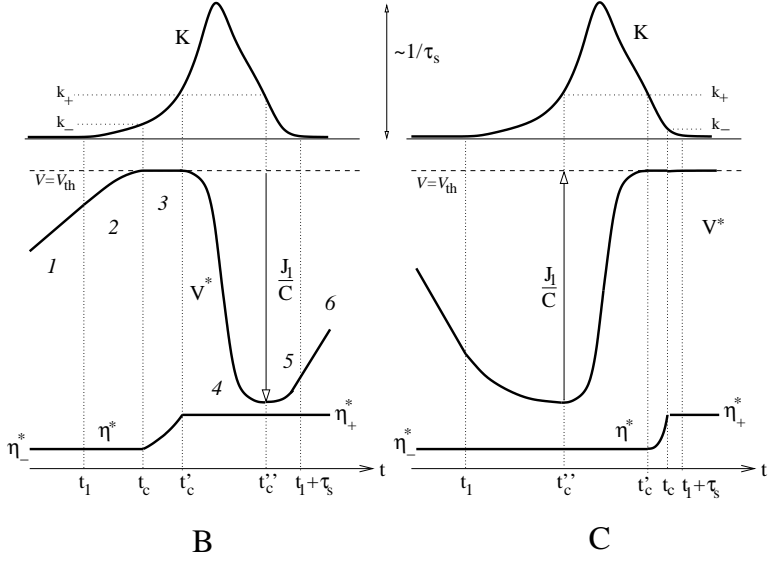,width=10cm}
\caption{Behaviors of the optimal potential $V^*$ (middle) and noise
$\eta ^*$ (bottom) close to a contact point, compared to the memory
kernel $K$ (top). An input of total amplitude $J_1$ enters the neuron
during the time interval $t_1<t<t_1+\tau_s$. Left: $J_1$ is strongly
negative as in Fig.~\ref{fig-sol1}B; italic numbers refer to the steps listed in the main text. Right: $J_1$ is strongly positive
as in Fig.~\ref{fig-sol1}C. See text for
a detailed description of the curves, of the constants $k_-,k_+$
(\ref{bess},\ref{bes}), and of the times $t_1,t_c,t'_c,t''_c,\tau
_s$. }
\label{fig-cont}
\end{center}
\end{figure}

To show that the noise always increases through an active contact, we 
consider that the synaptic integration is not instantaneous, but takes 
place over a finite albeit small time, $\tau_s$. We thus replace the 
expression for the current $I_i^{syn}$ in (\ref{defisyn}) with
\begin{equation}\label{defisyn2}
I_i^{syn}(t) = \sum _{j (\ne i)} J_{ij} \; \sum _{k} K( t - t_{j,k})
\end{equation}
where $J_{ij}$ is the strength of the connection from neuron $j$ onto 
neuron $i$, and $K(\tau)$ is is the memory kernel of the 
integration of synaptic entries
(top panel in Fig.~\ref{fig-cont}). We assume that $K(\tau)$ vanishes for
$\tau <0$ and for $\tau> \tau_s$ where the integration time $\tau_s$ 
is independent of the pair $(i,j)$. In addition, $K$ is positive, and
its integral over the interval $[0;\tau_s]$ is equal to unity.

We consider the case of a single incoming spike, as in Section 
\ref{illustration}. We want to show that, in the $\tau _s\to 0$
limit, the only constraint linking the values $\eta^*_-$ and 
$\eta^* _+$ of the optimal noise, 
respectively, before and after a spike entering at $t_1$, is
$\eta^*_+>\eta^*_-$, as we have found for a single incoming input 
in cases B and C. To do so, we assume that the
time of synaptic integration $\tau _s$ is small but {\em finite} , and
consider case B. The dynamics of $V^*$ and $\eta^*$ can be divided in several steps, whose numbered are reported on Fig.~\ref{fig-cont}:
\begin{enumerate}
\item  Prior to the input, {\em i.e.} at times $< t_1$, the
optimal noise $\eta^*_-$ is constant and the optimal potential $V^*$ is a linear 
function of the time, with slope $(I+\eta^*_-)/C$, as shown in Fig.~\ref{fig-cont}(left).
\item A strongly negative input of amplitude $J_1 (< J_-)$ is then received by 
the neuron between times $t_1$ and $t_1+\tau _s$. The derivative of
the potential now decreases with the time until
it vanishes at time $t_c$ defined through
\begin{equation}\label{bess}
K(t_c-t_1) = k_- \qquad \mbox{where} \qquad k_- \equiv
\frac{\eta ^*_-+I}{-J_1}\ .
\end{equation}
\item If the value of $\eta^*_-$ is chosen so that
$V^*(t_c)=V_{th}$, the potential tangentially reaches the
threshold at $t_c$ (contact point). Then, the potential remains
constant and equal to $V_{th}$. The noise obeys eqn. (\ref{eq2}) and, therefore, 
increases until it reaches the prescribed value, $\eta^*_+$, at time $t_c'$ such that
\begin{equation} \label{bes}
K(t'_c-t_1) = k_+ \qquad \mbox{where} \qquad k_+ \equiv
\frac{\eta ^*_++I}{-J_1}\ ,
\end{equation}
see bottom panel in Fig.~\ref{fig-cont}(left). 
\item Then the potential starts decreasing from its threshold value
through eqn. (\ref{optimb}), and reaches a minimum in $t''_c$, solution
of the same equation (\ref{bes}) as $t'_c$, see Fig.~\ref{fig-cont}(top left).
\item At later times the derivative of the potential is positive from
eqn. (\ref{optimb}), and increases until time $t_1+\tau_s$, coinciding with the end of the synaptic integration.
\item At times larger than $t_1+\tau_s$, the potential keeps growing with a
constant slope equal to $(I+\eta^*_+)/C$.
\end{enumerate}

In the $\tau_s \to 0$ limit, all times $t_c,t_c',t''_c$ tend to the
same value, that is, the time $t_1$. More precisely, as the slope of
$K$ is of the order of $\tau_s^{-2}$ (in absolute value), and
$\eta^*_-,\eta^*_+,V^*(t_1)$ are finite ($=O(1)$), then for $\tau_s\to
0$, $t_1,t_c,t'_c$ differ from each other by $O(\tau_s^2)$. Hence the
change in the potential $V^*$ between $t'_c$ and $t_1+\tau_s$ equals
$\frac{J_1}C+O(\tau_s)$. We conclude that, for $\tau_s\to 0$ the
potential becomes a discontinuous function of time with a
discontinuity $\frac{J_1}C$. In addition, the noise $\eta^*$ can also
jump abruptly from its value $\eta^*_-$ at $t_1^-$ to any {\em larger}
value $\eta^*_+$ at time $t_1^+$ since the maximum of $K$ tends to
infinity when $\tau_s\to 0$. 

Note that the drawing of Fig.~\ref{fig-cont}(left) tacitly assumes that $k_+>k_-$. 
A hypothetic scenario would be that the noise exactly compensates the synaptic input for a longer time interval (including the top of $K$), while the potential would remain equal to $V_{th}$. In this case, the peak value of the noise would be $O(1/\tau_s)$. The contribution to the integral (\ref{deflpcc}) would be of the order of $1/\tau_s$ and would diverge in the $\tau_s\to 0$ limit. Hence this possibility is precluded.

The above discussion is straightforwardly extended to case C. The optimal potential and noise are sketched in Fig.~\ref{fig-cont}(right). 
Note that the contact interval spreads beyond $[t'_c,t_c]$ in this
case. In the generic case of more than one incoming spikes, the contact
interval is restricted to $[t'_c;t_c]$ as in case B.
The noise can also discontinuously change from
its value $\eta_-^*<-I$ before the contact to any larger
value, $\eta^*_+$, after the contact.

\subsection{Case of $M\ge 2$ incoming spikes}\label{seccp2}

We now consider the general case of $M$
input spikes.  Let $V_0=0,m_0=1$ be, respectively, the initial value of the potential and the index of the first input spike. We define the piece-wise linear
function solution of (\ref{optimb}) for a constant noise $\eta$,
\begin{equation}\label{defvlin} V(\eta, t,t_0) = V_0 +\frac{I+\eta}C
\; (t-t_0) + \sum _{m=m_0}^M \frac{J_m}C\; \theta(t-t_m) \ .
\end{equation} 
We are looking for the smallest value of the noise
coefficient $\eta$ capable of bringing the potential $V(\eta,t,t_0)$ from
its initial value $V(\eta,t_0,t_0)=0$ to the threshold. The contact time, $t_c$, coincides with an entering spike, {\em i.e.}
$t_c=t_{m^*}$ for some $m^*\ge 1$. If $m^*=M+1$ then the optimal
potential is $V(\eta^*,t,t_0)$ throughout the inter-spike interval
$[t_0;t_{M+1}]$, and the problem is solved. If $m^*\le M$, $t_{m^*}$
is the first active contact point of the potential.  $\eta^*$ and
$V(\eta^*,t)$ are, respectively, the optimal noise and potential on
the interval $[t_0,t_{m^*}]$.

The correctness of the above statement can be established using a proof by 
contradiction.
\begin{itemize}
\item assume that the optimal noise, $\eta ^{opt}$, is smaller than $\eta^*$ on some 
sub-interval of $[t_0;t_c]$. Remark that the potential $V$ in (\ref{defvlin}) is an increasing function of the noise,
\begin{equation} \label{monoto}
\eta' > \eta \Longrightarrow V(\eta ',t,t_0) > V(\eta,t,t_0)
\ , 
\end{equation}
for all $t>t_0$. By virtue of (\ref{monoto}) and the definition of $\eta^*$, $V(\eta ^{opt},t,t_0)$ cannot touch the threshold at any time so the noise is constant throughout the interval $[t_0;t_c]$. Hence no active contact can take place at time $t_c$. As $\eta^*$ is the minimal value of the noise which can drive the potential into contact with the threshold over $[t_0;t_{M+1}]$, we conclude
that  $V(\eta ^{opt},t,t_0)$ cannot cross the threshold at any time $\le t_{M+1}$.
The neuron can therefore not spike at time $t_{M+1}$. 
\item conversely, suppose that the optimal noise is equal to $\eta^\alpha >\eta^*$ on the interval
$[t_0;t_{m^\alpha}]$ with $1\le m^\alpha< m^*$, and takes another value on the interval
$[t_{m^\alpha};t_c]$\footnote{The case of three or a higher number of values for the noise
can be handled exactly in the same way.}. 
As the change of noise can take place only through an active contact,
and the change is necessarily positive (Section \ref{seccp-sec}), we have $\eta^\beta > \eta^\alpha$.
Applying  (\ref{monoto}) to the interval $[t_{m^\alpha};t_c]$, we have 
\begin{equation}
V(\eta^\beta ,t_c,t_{m^\alpha}) > V(\eta^\alpha,t_c,t_{m^\alpha})
\ . 
\end{equation}
Adding the value of the optimal potential in $t_{m^\alpha}$ to both members of the 
previous inequality, we find
\begin{eqnarray}\label{lp}
V^*(t_c) &=& V(\eta^\beta ,t_c,t_{m^\alpha})+V(\eta^\alpha,t_{m^\alpha},t_0) 
\nonumber \\ &>&  V(\eta^\alpha,t_c,t_{m^\alpha})+V(\eta^\alpha,t_{m^\alpha},t_0) \nonumber \\
&=& V(\eta^\alpha,t_c,t_0) \nonumber\\ &>& V(\eta^*,t_c,t_0) 
\end{eqnarray}
where the last inequality comes from (\ref{monoto}). But, by definition of $\eta^*$,
$V( \eta^*,t_c,t_0) =V_{th}$. Hence, we find that the optimal potential in $t_c$ is above threshold,
which cannot be true.
\end{itemize}

The optimal noise and potential on the
remaining part $[t_c;t_{M+1}]$ of the inter-spike interval can be determined
iteratively. We replace $t_0$ with $t_{m^*}$ and $V_0$ with $V_{th}$
if $J_{m^*}>0$ or $V_{th}+\frac{J_{m^*}}C$ if $J_{m^*}<0$ in
(\ref{defvlin}), and look for the lowest noise producing a new contact
point over the interval $[t_{m^*}, t_{M+1}]$. The procedure is
repeated until the whole interval is exhausted. This way an increasing sequence
of noise values is obtained, each corresponding to the
slope of the optimal potential between two successive contact points.

\section{Passive contacts}
\label{secgnon}

When the membrane leaking conductance is non zero, some change have
to be brought to the above calculation of the optimal noise and
potential. First, in the absence of inputs, the noise is no longer constant, 
but rather it is an exponentially increasing (in absolute value) 
function of the time (\ref{eq1}).
Similarly, the potential $V^*$ itself is not a linear function of the time as in
(\ref{lp}), but is a linear
combination of $\exp(\pm t/\tau)$ with appropriate coefficients, see (\ref{defv}).

The main conclusion of Appendix \ref{seccp} still holds: the difference
between the noise values just after and before an active contact
point, coinciding with a synaptic input, is always positive (Fig. ~\ref{fig-scheme}A). 
Consequently, the procedure of Section \ref{sec-algo}, {\em i.e.}
the iterative search for the active contact points and the minimal
noise coefficient $\eta ^*$, defined through (\ref{defnoise}), remains unchanged.
Note that some care must be taken to translate the statement about the growth of the noise to the values of the noise coefficients. 
Consider for instance two successive contact times, $t$ and $t'$, and call $\eta$, 
$\eta'$ the corresponding noise coefficients. That the noise is larger at time $t'$ than at time $t$ implies that $\eta \times \exp ((t'-t )/\tau) < \eta'$, but does not imply that $\eta'$ is larger than $\eta$ \footnote{This situation can not happen in the $g=0$ case, where the noise and the noise coefficient coincide.}.

\begin{figure}
\begin{center}
\epsfig{file=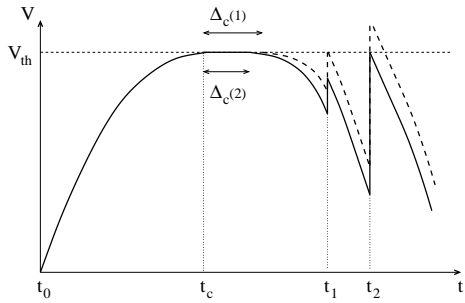,width=6cm}
\caption{Sketch of the optimal potential close to a passive contact starting at time $t_c$. The duration of the passive contact is $\Delta _c(\ell)$, where $\ell$ is the index of time $t_\ell$ corresponding to the next active contact. The potentials corresponding to the
hypothesis $\ell=1$ and $\ell=2$ are shown with the dashed and full curves respectively.}
\label{passc}
\end{center}
\end{figure}

There exists, however, a major 
difference between the $g=0$ and $g\ne 0$ cases. When $g>0$, the optimal potential 
is not guaranteed to be a monotonous function of the time, as shown in Fig.~\ref{passc}. 
For given values of $g,I$, and of the times and the amplitudes of the synaptic inputs,
the optimal potential $V^*$ may touch the threshold at an intermediate time, $t_c$.
Such a situation is called passive contact.
It is important to note that the value of the optimal noise during a 
passive contact is, according to eqn. (\ref{eq2}), equal to $g V_{th}-I$.
As the optimal noise is a monotonous function of the time between two active
contacts, see eqn (\ref{eq1}), 
the value $g V_{th}-I$ can be crossed at most once: there is at most one
passive contact in between two successive active ones. To be more precise,  there are at most $A+1$ passive contacts in an inter-spike interval with $A$ active contacts.

To decide the existence of a passive contact in an interval $[t_0;t_{M+1}]$, we look for a solution of the two coupled equations expressing that the optimal potential touches the threshold without crossing it,
\begin{equation}
V^*(\eta_{p}, t_{c})=V_{th} \quad \hbox{\rm and} \quad
\frac{\partial V^*}{\partial t} (\eta, t_{c})=0 \ .
\end{equation}
The solutions of these equations give the noise coefficient $\eta_{p}$ and the
contact time $t_c$ at which the optimal potential reaches the threshold value (Fig.~\ref{passc}). The solution can be calculated analytically, with the following result.
Let us call $V_0$ the value of the potential of the neuron at time $t_0^+$. For each $m\le M$ we define 
\begin{equation}
V_m = V_0 +\sum _{\ell: t_{0}<t_\ell\le t_{m}} \frac{J_\ell}C \;e^{( t_\ell-t_0)/\tau}\ ,
\end{equation}
where the summation runs overs the spikes entering the neuron between
times $t_0$ and $t_c$. A passive contact takes place in the interval $[t_{m};t_{m+1}]$ if:
\begin{description}
\item[$\bullet$] $g V_{th}-I$ and $V_m-V_{th}$ have the same sign;
\item[$\bullet$] the noise coefficient 
\begin{equation}\label{etap}
\eta _p =gV_m-I-\sqrt{(gV_m-I)^2-(gV_{th}-I)^2}
\end{equation}
is smaller than the lowest noise coefficient corresponding to all the possible active contacts at times $t_{\ell}$, with $1\le \ell \le M$;
\item[$\bullet$] the corresponding contact time
\begin{equation}\label{etap2}
t_{c}= t_{0}-\tau \; \log\left[ \frac{\eta _p}{gV_{th}-I}\right] \ ,
\end{equation}
where $\eta _{p}$ is given by (\ref{etap}), lies in the correct interval: $t_{m}< t_c< t_{m+1}$;
\item[$\bullet$] the optimal potential can reach again the threshold at a later time, coinciding with an input spike or with the end of the inter-spike interval. We call $\Delta _c(\ell)$ the duration of the active contact such that  the potential reaches $V_{th}$ at time $t_\ell$, starting from $V_{th}$ at  time $t_c+\Delta_{c}(\ell)$, see Fig.~\ref{passc}. 
The analytical expression for the duration of the passive contact allowing the potential to be in active contact at time $t_\ell$ is
\begin{equation}\label{eqdeltac}
\Delta_c (\ell) = -\tau \log \left\{\frac 1{2\, V_{a}(\ell)} \bigg[ V_{b}(\ell) - \sqrt{
V_{b}(\ell)^2 - \left(V_{th}-\frac Ig \right)^2}\ \bigg] \right\}\ .
\end{equation}
where
\begin{equation}
V_{a}(\ell) = \frac{\eta_p}{2g}\; e^{(t_\ell-t_0)/\tau}\quad \hbox{\rm and}\quad
V_{b}(\ell) = V_{th}
- \frac Ig - \sum _{\ell'<\ell} \frac{J_{\ell'}}C \;e^{-(t_\ell-t_{\ell'})/\tau}
-\frac {J_\ell}C\;\theta(J_\ell) \ .
\end{equation}
We must have $t_{c}+\Delta_{c}(\ell)<t_{\ell}$ for at least one value of $\ell\ge m+1$.
\end{description} 
When all the conditions are fulfilled, a passive contact is present. The duration of the contact, $\Delta_c$, merely plays the role of a latency time after which the potential $V^*$ resumes its course (Fig~\ref{passc}). We can check that $V^*$ is an increasing function of $\Delta _c$.
The optimal value of $\Delta _c$ will therefore be equal to the smallest possible value of $\Delta_{c}(\ell)$, for the very same reason that we had to chose the minimal noise when looking for active 
contacts, see example in Fig.~\ref{passc}. 

To end this Appendix, we give the expression for the
log-likelihood $L^*$ (\ref{deflpcc}) for an interval including a passive
contact between two active contacts. Gathering the contributions to the integral
of the squared optimal noise coming from the three intervals $[t_0;t_c]$, 
$[t_c, t_c+\Delta _c]$, and $[t_c+\Delta _c;t_{m^*}]$, we obtain
\begin{equation}
L^*({\cal T}|{\cal J},{\cal I})= -\frac{(gV_{th}-I)^2}2 \left\{ \Delta _c +
\tau\; \frac{\exp\big[2( t_{m^*}-t_c-\Delta _c)/\tau\big] -
\exp \big[-2(t_c-t_0)/\tau\big] 
}{2}\right\}\ .
\end{equation}
Differentiation of $L^*$ with respect to the current and couplings gives the
expressions for the gradient and Hessian matrix needed for the Newton-Raphson
method. The expressions are easy to obtain but are lengthy, and thus we do not
reproduce them.

\section{On the eigenmodes of the Hessian matrix for weak couplings}
\label{aeigen}

In this Appendix, we analyze the eigenmodes and eigenvalues of the Hessian matrix of the log-likelihood $L^*$, and relate the eigenmodes to the fluctuations of the effective current, $I_i^e$, of the current, $I_i$, and of the couplings, $J_{ij}$.
Consider two successive spikes emitted by neuron $i$ and the optimal
potential $V^*_i(t)$ on the time interval $[t_{i,k}; t_{i,k+1}]$.
When the couplings $J_{ij}$ vanish and passive contacts are absent, $V_i^*(t)$ does not enter into contact with the threshold at times $<t_{i,k+1}$. 
By continuity, this statement remains true if the couplings $J_{ij}$ 
have very small amplitudes. In this regime, the stochastic process undergone
by the potential is simply the Ornstein-Uhlenbeck process with a time-varying force,
and the expression for $L^*$ (\ref{deflpcc}) is exactly given by
\begin{equation}
L^{*} ({\cal T}|{\cal J},{\cal I}) = - \frac 12\sum _{i,k}  {\mu} ^{(i)}_{k}\;
\bigg(C\; V_{th}-\sum _{j (\ne i)}J_{ij} \; \phi^{(i)}_{k,j} -I_i\,\tau \;
\phi^{(i)}_{k,i}\bigg)^2
\end{equation}
where 
\begin{equation}
{\mu} ^{(i)}_{k} =  \frac2{\tau} \; \bigg(1-e^{-2(t_{i,k+1}-t_{i,k})
/\tau}\bigg)^{-1} \ ,
\end{equation} 
and
\begin{equation}
\phi^{(i)}_{k,j} = \left\{ \begin{array} {c c c}\displaystyle{
\sum _{l}  e^{-(t_{i,k+1}-t_{j,l})/\tau} \; \theta \big( t_{i,k} < t_{j,l} < t_{i,k+1} \big)} &
\hbox{\rm if} & j \ne i \ , \\
1- e^{-(t_{i,k+1}-t_{i,k})/\tau}&
\hbox{\rm if} & j = i \ .\end{array} \right.
\end{equation}
The Hessian matrix of $L^{*}$, attached to neuron $i$, 
is the $N\times N$ matrix (\ref{fluch}) with elements
\begin{equation}\label{fluch2}
\sigma^2 {\bf H} ^{(i)}_{jj'} = \sum _k {\mu} ^{(i)}_{k}\; \phi^{(i)}_{k,j}\;\phi^{(i)}_{k,j'} \ ,
\end{equation}
${\bf H}^{(i)}$ is a positive matrix according to (\ref{fluch2}). To study its spectrum let
us first consider the case of very weak leakage 
(very large $\tau$). In this limit, calling
$\Delta t_k^{(i)} = t_{i,k+1}-t_{i,k}$ the duration of the $k^{th}$ ISI of neuron $i$,
we have
\begin{equation}
{\mu} ^{(i)}_{k} \to \frac 1{\Delta t_k^{(i)}}\ , \quad
\phi^{(i)}_{k,i} \to \frac {\Delta t_k^{(i)}}\tau \ ,\quad 
\phi^{(i)}_{k,j} \to \hbox{\rm nb. of spikes of neuron $j$ 
in the}\ k^{th}\ \hbox{\rm ISI of neuron $i$.}
\end{equation}
Let us define the firing rate $f^{(i)}_{k,j}$ of neuron $j (\ne i)$ 
in the $k^{th}$ ISI of neuron $i$, and $f^{(i)}_{ki}=\frac 1\tau$. We obtain
\begin{equation}\label{fluch3}
\frac {\sigma^2}T\; 
{\bf H} ^{(i)}_{jj'} =\frac 1T \sum _k \Delta t_k^{(i)}\; f^{(i)}_{k,j}\;f^{(i)}_{k,j'} \ .
\end{equation}
where $T$ is the duration of the recording. The right hand side of the above equation
can be interpreted as the covariance matrix of the rates $f^{(i)}_{k,j}$, where each
ISI of neuron $i$ is weighted proportionally to its duration. For vanishing couplings,
these instantaneous rates  are decoupled from neuron to neuron. Hence, 
$f_{k,j}^{(i)}$ fluctuates around the average firing rate $f_j$ 
(number of the spikes fired by neuron $j$, divided by $T$),
with a variance we denote by $\langle f_j ^2\rangle_c$. This statement
holds for $j\ne i$; in addition we define  
$f_i = \frac 1\tau$. Neglecting terms of the order of $\tau^{-2}$,
we end up with the following approximation for the Hessian matrix,
\begin{equation}\label{fluch4}
\frac {\sigma^2}T\; {\bf H} ^{(i)}_{jj'} = f_j\; f_{j'} + 
\delta _{j,j'}\; \omega_j \qquad \hbox{\rm where} \qquad 
\omega _j = \left\{\begin{array}{c c c}
\langle  f_j ^2\rangle_c &\hbox{\rm if} & j \ne i \ ,\\
0 &\hbox{\rm if} & j = i \ . \end{array}\right. \ ,
\end{equation}
which becomes exact in the limit of infinitely long recordings.

The matrix ${\bf H}^{(i)}$ is the sum of a rank one matrix plus a diagonal
matrix. For small values of $\sigma$, the fluctuations of the firing rates, represented by 
the $\omega _j$'s, are expected to be small compared to the product of any two 
average firing rates. We immediately deduce that the largest eigenvector of the matrix,
${\bf v}_{max}$, has components 
$({\bf v}_{max})_j= f_j$ for all $j=1,\ldots, N$. The associated
eigenvalue, $\lambda_{max}$, is given by
\begin{equation}
\frac{\sigma^2}S\;\lambda _{max} = \frac TS\; \sum _{j }   (f_j)^2  \ ,
\end{equation}
where $S$ is the total number of spikes.
If the neurons have quantitatively similar firing rates $\simeq \langle f\rangle$, 
then $\sigma^2 \lambda_{max}/S \simeq \langle f\rangle$.
The probability density of vector $v^{(i)}$ is
\begin{equation} \label{fluc}
P (\{v^{(i)}\} |{\cal T}) \simeq P (\{\hat v^{(i)}\}|{\cal T}) 
\times \exp \left[ - \frac 1{2} \;\sum_{i,j,j'} 
\big(v^{(i)}_j - \hat v^{(i)}_j\big) {\bf H}  ^{(i)}_{j,j'}
 \big(v^{(i)}_{j'} - \hat v^{(i)}_{j'}\big) \right] \ .
\end{equation}
A fluctuation $\delta {\bf v}= \{\delta (I_i\tau), \delta J_{ij}\}$ around the most likely
values for the current and the couplings, and along vector ${\bf v}_{max}$, will 
change the log-likelihood by
\begin{equation}
\delta \bigg(\frac{\log P}S\bigg)=-\frac{\lambda _{max}\; 
(\delta {\bf v} \cdot {\bf v}_{max})^2}{2\; S\;({\bf v}_{max})^2}
= -\frac {T}{2\,\sigma^2\;S}\; \bigg(\delta I_i +\sum _{j (\ne i)} J_{ij}\, f_j\bigg)^2
\simeq-\frac{(\delta I_i^e)^2}{2\,\sigma^2\,N\,\langle f\rangle}
\ .
\end{equation}
Hence the effective current $I_i^e$ is associated to the largest eigenmode, and is the
parameter requiring the least number of data to be inferred.

We now look for the smallest eigenvalue, $\lambda _{min}$. 
Numerical investigations suggest that the associated eigenvector,  ${\bf v}_{min}$, 
correspond to fluctuations of the current $I_i$ only. 
We thus assume that the components of ${\bf v}_{min}$ are: $({\bf v}_{min})_i=1$, and
$({\bf v}_{min})_j=-\epsilon _j$ with $\epsilon_j \ll 1$.
The eigensystem we need to solve is
\begin{eqnarray}
\sigma^2\,\lambda _{min} &=& \frac T\tau \bigg(\frac 1\tau - \sum _{j' (\ne i)}
\epsilon _{j'}f_{j'}\bigg) \label{eqt1}\\
-\sigma^2\,\lambda _{min}\; \epsilon_j &=& T\;f_j\; \bigg(\frac 1\tau - \sum _{j' (\ne i)}
\epsilon _{j'}f_{j'}\bigg) -T\;\omega_j \; \epsilon_j \qquad \forall\ j (\ne i)
\ .\label{eqt2}
\end{eqnarray}
According to (\ref{eqt2}) and (\ref{eqt1}), we have, for all $j (\ne i)$,
\begin{equation}\label{epsj7}
\epsilon _j = \frac{f_j\;\tau\;\sigma^2\;
\lambda_{min}}{\omega_j\;T} + O(\sigma^2\;\lambda_{min}\,\epsilon_j)\ .
\end{equation}
Inserting this expression for the components $\epsilon_j$ 
of the eigenvector into (\ref{eqt1}), we obtain
\begin{equation}
\frac{\sigma^2}S\;\lambda _{min} = \frac T{
\displaystyle{\tau^2\; \bigg(1 + \sum _{j (\ne i)} \frac{f_j ^{\, 2}}{\omega_j}}
\bigg)\; S}\ .
\end{equation}
If all neurons have quantitatively similar firing rates, $\langle f\rangle$, and 
variances,  $\langle f^2\rangle_c$, we obtain 
$\sigma^2\lambda_{min}/S\simeq \langle f^2\rangle_c/( N^2\langle f\rangle^3\tau^2)$. According to 
(\ref{epsj7}), the components $\epsilon _j$ of the eigenvector are very small,
 $\epsilon _j \simeq 1/(N^2\tau\langle f\rangle)$, for all $j\ne i$. 
Hence ${\bf v}_{min}$
is localized on its current component only. A fluctuation 
$\delta {\bf v}= \{\delta (I_i\tau), \delta J_{ij}\}$, where the $\delta J_{ij}$'s are chosen
to be orthogonal to all the other eigenmodes of ${\bf H}^{(i)}$, 
modifies the log-likelihood by
\begin{equation}
\delta \bigg(\frac{\log P}S\bigg)=-\frac{\lambda _{min}\; 
(\delta {\bf v} \cdot {\bf v}_{min})^2}{2\; S\;({\bf v}_{min})^2}
\simeq -\frac {  \langle f^2\rangle_c\; \big(\delta I_i\big)^2}{
2\,\sigma^2\;N^2\;\langle f\rangle^3}
\ .
\end{equation}
We conclude that the current $I_i$ 
is the hardest parameter to infer, {\em i.e.} the one requiring the largest number of data.

When the membrane leaking time becomes of the order of, or smaller than the 
average ISI duration, the above calculation has to be modified. 
From a qualitative point of view, the average firing rate $f_{j}$ must now be
defined as the mean number of spikes emitted by the neuron $j$ in a time-window
of duration $\tau$ preceding a spike of neuron $i$, divided by $\tau$, see (\ref{deffjitau}). 
The eigenvector of ${\bf H}^{(i)}$ with largest eigenvalue $\lambda_{max}$ is still 
given by $({\bf v}_{max})_j=f_j$, with $f_i=1/\tau$, and 
\begin{equation}
\frac{\sigma^2}S\;\lambda _{max} \simeq \frac \tau{N}\; \sum _{j}   (f_j)^2  \ .
\end{equation}
Again, these fluctuations are associated to the effective current, with  the 
newly defined average firing rates $f_i$. As $\tau$ gets smaller and smaller, 
all the rates $f_j$ with $j\ne i$ become smaller and smaller compared to $f_i$, 
and the effective current $I_i^e$ gets closer and closer to the true current $I_i$. 
Obviously, the inference of the synaptic coupling $J_{ij}$ is possible if the
firing rate $f_j$ defined on a time-window of duration $\tau$ preceding a spike 
of neuron $i$ is much larger than $1/T$.

\section{Fluctuations of the potential around the optimal path at small noise}
\label{app-corrections}

In this Appendix, we derive formula (\ref{result-fluctu}) for the fluctuations of the potential around its optimal value at the mid-point of the ISI.
A useful formulation for  $p_{FPT}$ in (\ref{prodfpt})
can be given in terms of a path integral over the potential,
\begin{eqnarray}\label{pathint}
 p_{FPT} (t_{i,k+1} &|& t_{i,k} , \{t_{j,l} \}, \{J _{ij}\}, I_i )= \\
&&- \frac{\partial}{\partial t_{i,k+1}}\,\int _{V_i(t_{i,k}^+)=0}
^{V_i(t_{i,k+1}^-)<V_{th}} {\cal D}V_i(t)\; \exp \left( -\frac 1{2\sigma^2}
\ {\cal L}[ V_i(t); k,{\cal T},{\cal J},{\cal I}]\right) \ . \nonumber
\end{eqnarray}
The measure ${\cal D}V_i(t)$ in the path-integral (\ref{pathint}) 
is restricted to
the potentials $V_i(t)$ remaining smaller than the threshold $V_{th}$ at all 
times $t$. The upper bound $V_i(t_{i,k+1}^-)<V_{th}$ means that the integral is performed
over all the values of the potential smaller than $V_{th}$ at time $t_{i,k+1}^-$,
while $V_i(t_{i,k})$ is constrained to be zero.

We introduce the dimensionless variable 
$\psi_i(t)=(V_i(t)-V_i^*(t))/V_{th}$ to represent the time-dependent 
fluctuation of the potential (Fig.~\ref{fig-fluctuphi}C).
According to (\ref{pathint}) and (\ref{pathint2}), 
the log probability density of a path-fluctuation $\psi_i(t)$ on the
inter-spike interval $[t_{i,k};t_{i,k+1}]$ is, after multiplication by $\sigma^2$,
\begin{eqnarray}\label{pathint3}
&& {\cal L}[\psi_i(t)\,V_{th}+V^*_i(t); k,{\cal T},{\cal J},{\cal I}]
\nonumber \\ &=& {\cal L}[V^*_i(t) ; k,{\cal T},{\cal J},{\cal I}]
+ \frac{V_{th}^2}{2} \int _{t_{i,k}}^ {t_{i,k+1}}
dt'\;  \int _{t_{i,k}}^ {t_{i,k+1}}dt\; \psi_i(t') \; 
\frac{\delta ^2{\cal L}}{\delta V_i^*(t') \,\delta V_i^*(t) } \;
\psi_i(t) + O(\psi _i^3) \nonumber \\
&=& L^*({\cal T}|{\cal J},{\cal I}) - \frac{V_{th}^2}{2} \int _{t_{i,k}}^ {t_{i,k+1}}
dt\; \psi_i(t)  \bigg[ -C^2 \frac{d^2}{dt^2} +g^2 \bigg] \psi_i(t) + O(\psi _i^3) \ ,
\end{eqnarray}
up to an additive term independent of $\psi_i$.
Note that we have used the optimality condition (\ref{optima})
to exclude terms linear in $\psi_i$ in (\ref{pathint3}). 
We now want to perform the path integral over the fluctuations
$\psi_i(t)$ in (\ref{pathint}). When $\sigma$ is small we may discard the cubic and higher order terms in $\psi_i$. The boundary condition on $\psi_i$ are
$\psi_i(t_{i,k})=\psi_i (t_{i,k+1})=0$: the values of the potential $V_i(t)$ 
are constrained right after and before the emission of a spike, and, hence,
cannot fluctuate (Fig.~\ref{fig-fluctuphi}C). We therefore write the fluctuations $\psi_i(t)$ as the following Fourier series,
\begin{equation}
\psi_i (t) = \sum _{n\ge 1} \psi _n \; \sin \left( \frac {n\,\pi\, (t-t_{i,k})}{t_{i,k+1}-t_{i,k}}\right)
\ ,
\end{equation}
where the $\psi_n$ are stochastic coefficients. The integral on the last line of (\ref{pathint3}) can be calculated with the result
\begin{equation}\label{pathint4}
\frac{V_{th}^2}{2} \int _{t_{i,k}}^ {t_{i,k+1}}
dt\; \psi_i(t)  \bigg[ -C^2 \frac{d^2}{dt^2} +g^2 \bigg] \psi_i(t) 
=\frac{\rho\,(CV_{th})^2}{4\tau}\sum _{n\ge 1}\big[1+\left( \frac{n\pi} \rho\right)^2\big]\,  \psi_n^2 
\ ,
\end{equation}
where 
\begin{equation}
\rho=\frac{t_{i,k+1}-t_{i,k}}\tau 
\end{equation}
is the duration of the ISI measured in units of the membrane leaking time.
Hence, if we relax the constraint that the fluctuating potential should remain below threshold at all times, the $\psi_n$'s are independent Gaussian variables with zero means and variances 
\begin{equation}
\lambda_n = \frac{2\, \tau\, \sigma ^2}{(CV_{th})^2}\, \frac {\rho}{\rho^2+n^2\pi^2}= 
 \frac {2\,\bar\sigma ^2\, \rho}{\rho^2+n^2\pi^2}\ ,
\end{equation}
where $\bar\sigma$ is defined in (\ref{defnounit}). We may now calculate the variance of $\psi_i$ at the mid-point of the ISI, see Fig.~\ref{fig-fluctuphi}C,
\begin{equation} \label{pathint5}
\left\langle \psi_i \left( \frac{t_{i,k}+t_{i,k+1}}2\right)^2 \right\rangle=\left\langle \left[ \sum _{n\ge 1} \psi_n \, \sin \left(\frac{n\pi}2\right)\right]^2\right> = \sum _{p\ge 0} \lambda_{2p+1} \ .
\end{equation} 
Summing up the series over $p$ in (\ref{pathint5}) gives
expression (\ref{result-fluctu}).

\section{Expression of the moving threshold and alternative procedures}
\label{essai2}

In Section \ref{sec-beyond} we explain that the value of the moving threshold, $V_{th}^M$, is estimated from the intersection of the tangent to the probability of survival in $V=V_{th}$ with the $p_s=\frac 12$ line. Hence,
\begin{equation}
V_{th}^M = V_{th}   + \left(2\, \frac {d p_s}{dV}(\delta t|V=V_{th}) \right)^{-1}
\ .
\end{equation}
The slope of $p_s$ can be expressed in terms of a series of parabolic cylinder functions (Alili, Patie and Perdersen, 2005),
\begin{equation}\label{expressfpt}
\frac {d p_s}{dV}(\delta t|V=V_{th}) = - 
\sum _{i\ge 0} \frac {\exp(-n_i \, \delta t/\tau)}{n_i\,L_i} \; 
D'_{n_i} \left( \frac{\sqrt{2gC}}\sigma \;\big (\frac Ig-V_{th}\big)\right)^2
\ ,
\end{equation}
where $D'_n(z)$ denotes the derivative of Weber's function of order $n$, $D_n(z)$, with respect to its argument $z$. The normalization coefficients are
\begin{equation}
L_{i} = \int _{-\infty}^{V_{th}} dV\;  
D_{n_i} \left( \frac{\sqrt{2gC}}\sigma \;\big (\frac Ig-V\big)\right)^2
\ .
\end{equation}
The orders $n_i$, $i=0,1,2,\ldots$, are the roots of the equation (Mei and Lee, 1983) 
\begin{equation}
D_n \left( \frac{\sqrt{2gC}}\sigma \;\big (\frac Ig-V_{th}\big)\right) = 0 
\end{equation} 
with $0 < n_0 < n_1 < n_2 < \ldots$. The gap between successive levels, $n_{i+1}-n_i$, is larger than 1. Note that the contributions from high orders $n_i$ decay exponentially with $\delta t/\tau$ in (\ref{expressfpt}). Hence, in practice, the summation can be carried out over a finite number of terms.

\begin{figure}
\begin{center}
\epsfig{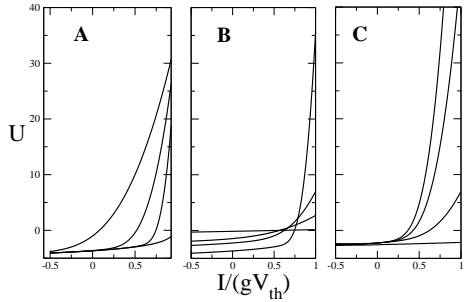}
\caption{Cost-energy function $U$ over the current
as a function of the ratio $I/(gV_{th})$ for different values of 
$\sigma$ ({\bf A}),
$g$ ({\bf B}), and the inter-spike interval $\delta t$ ({\bf C}). Values
of the parameters are: 
{\bf A.} $\delta t/\tau=.025$, and $\sigma/(V_{th}\sqrt{gC})
=.016,.16,.32,.64$ from right to left; 
{\bf B.} $g=1,5,10,40 \; C/\delta t$ 
from top to down on the left side, with $\sigma/\sqrt{ \delta t}=I$; 
{\bf C.} $\sigma/(V_{th}\sqrt{gC})=.32$,
 and $\delta t/\tau=1,10,50,100$ from bottom to up.}
\label{fig-4}
\end{center}
\end{figure}

The Moving Threshold procedure was designed to take into account the effects of a moderate noise level, $\sigma$. An alternative approximate procedure consists in subtracting to the log-likelihood a cost-function preventing the current, or the effective current from getting too close to $gV_{th}$. For a quantitative treatment consider a single neuron in the absence of synaptic input, for which $p_{FPT}$ can be calculated under the form of a series of parabolic cylinder functions, see above. We denote by $p_{FPT}^{cl}$ the approximation to $p_{FPT}$ obtained when taking into account the optimal path only.
We define the cost-energy function
\begin{equation}
U(I;g,\sigma,\tau)=
\log \left[ \frac{p_{FPT} (\delta t ;g,\sigma,I)}
{p_{FPT}^{cl} (\delta t;g,\sigma,I)}\right]
\ .
\end{equation} 
for the current $I$. We show in Fig.~\ref{fig-4} the shape of $U$ for different values of $g$, $\sigma$, and the inter-spike interval $\delta t$. As expected from above, this cost function is essentially flat when $I/(gV_{th})\ll 1$, and is repulsive when 
$I/(g V_{th}) \to 1$. The repulsion is strong when the inter-spike interval, $\delta t$, the membrane conductance, $g$, and the noise standard deviation, $\sigma$, are large. 

In presence of synaptic inputs, we approximate the non-perturbative corrections by subtracting $(N_i-1)\; U(I_i^e)$ to our log-likelihood, where $N_i$ is the number of spikes of neuron $i$, and $I_i^e$ its effective current. This simple approximation preserves the concavity of the log-likelihood and is computationally  simple since $U$ has to be calculated only once for each step and neuron.  Simulations show that the performance of the inference algorithm with the cost function $U$ is quantitatively similar to the one obtained with the Moving Threshold procedure.

\end{document}